\documentclass[11pt]{article}

\usepackage[final]{acl}

\usepackage{times}
\usepackage{latexsym}
\usepackage{booktabs}
\usepackage{amsmath}
\usepackage{multirow}
\usepackage{xcolor}
\usepackage{colortbl}
\usepackage[T1]{fontenc}

\usepackage[utf8]{inputenc}

\usepackage{microtype}

\usepackage{inconsolata}
\usepackage{subcaption}
\usepackage{graphicx}

\title{When Agents Look the Same: Quantifying Distillation-Induced Similarity in Tool-Use Behaviors}

\author{
  \textbf{Chenghao Yang\textsuperscript{1,2}},
  \textbf{Yuning Zhang\textsuperscript{1,2}},
  \textbf{Zhoufutu Wen\textsuperscript{3,$\dagger$}},
  \textbf{Tao Gong\textsuperscript{1,2,$\dagger$}},
\\
  \textbf{Jiaheng Liu\textsuperscript{4}},
  \textbf{Qi Chu\textsuperscript{1,2}},
  \textbf{Nenghai Yu\textsuperscript{1,2}}
\\
\\
  \textsuperscript{1}School of Cyber Science and Technology, USTC,
\\
  \textsuperscript{2}Anhui Province Key Laboratory of Digital Security,
  \textsuperscript{3}M-A-P, 
  \textsuperscript{4}NJU
\\
\small{\texttt{yangchenghao@mail.ustc.edu.cn}}, \texttt{tgong@ustc.edu.cn}, \texttt{wzft123@outlook.com}
}

\usepackage[normalem]{ulem}
\definecolor{colorAnthropic}{HTML}{e9e9e9}  
\definecolor{colorOpenAI}{HTML}{F8FFF8}     
\definecolor{colorDeepSeek}{HTML}{F5F8FF}   
\definecolor{colorMoonshot}{HTML}{FFF8FA}   
\definecolor{colorByteDance}{HTML}{FFFCF5}  
\definecolor{colorGoogle}{HTML}{F8FFFC}     
\definecolor{colorQwen}{HTML}{FFFBF9}  
\definecolor{colorZhipu}{HTML}{FAFAFF}      
\usepackage[most]{tcolorbox}
\usepackage{enumitem} 
\usepackage{amssymb} 
\usepackage{algorithm}
\usepackage{algorithmic}
\usepackage{listings}
\usepackage[most]{tcolorbox}
\tcbuselibrary{listings}
\usepackage{float}

\newtcblisting[auto counter, number within=section]{Prompt}[2][]{%
  colback=white,
  colframe=cyan,
  width=\linewidth,  
  arc=5mm,
  boxrule=0.8mm,
  title=\large #2,
  breakable,
  fonttitle=\small,
  listing only,  
  listing options={
    basicstyle=\footnotesize\ttfamily,
    breaklines=true,  
    breakautoindent=false,
    columns=flexible,
  },
  #1
}
\tcbuselibrary{breakable}
\begin{document}
\maketitle
{
  \let\thefootnote\relax
  \footnotetext{$\dagger$ Corresponding authors.}
}
\begin{abstract}
Model distillation is a primary driver behind the rapid progress of LLM agents, yet it often leads to behavioral homogenization. 
Many emerging agents share nearly identical reasoning steps and failure modes, suggesting they may be distilled echoes of a few dominant teachers. 
Existing metrics, however, fail to distinguish mandatory behaviors required for task success from non-mandatory patterns that
reflect a model's autonomous preferences. 
We propose two complementary metrics to isolate non-mandatory behavioral patterns: \textbf{Response Pattern Similarity (RPS)} for verbal alignment and \textbf{Action Graph Similarity (AGS)} for tool-use habits modeled as directed graphs. 
Evaluating 18 models from 8 providers on $\tau$-Bench and $\tau^2$-Bench against Claude Sonnet 4.5 (thinking), we find that within-family model pairs score 5.9 pp higher in AGS than cross-family pairs, and that Kimi-K2 (thinking) reaches 82.6\% $S_{\text{node}}$ and 94.7\% $S_{\text{dep}}$, exceeding Anthropic's own Opus 4.1. 
A controlled distillation experiment further confirms that AGS distinguishes teacher-specific convergence from general improvement. 
RPS and AGS capture distinct behavioral dimensions (Pearson $r$ = 0.491), providing complementary diagnostic signals for behavioral convergence in the agent ecosystem.
Our code is available at \url{https://github.com/Syuchin/AgentEcho}.
\end{abstract}


\begin{figure}[!t]
\centering
\begin{subfigure}[b]{\columnwidth}
  \centering
  \includegraphics[width=\columnwidth]{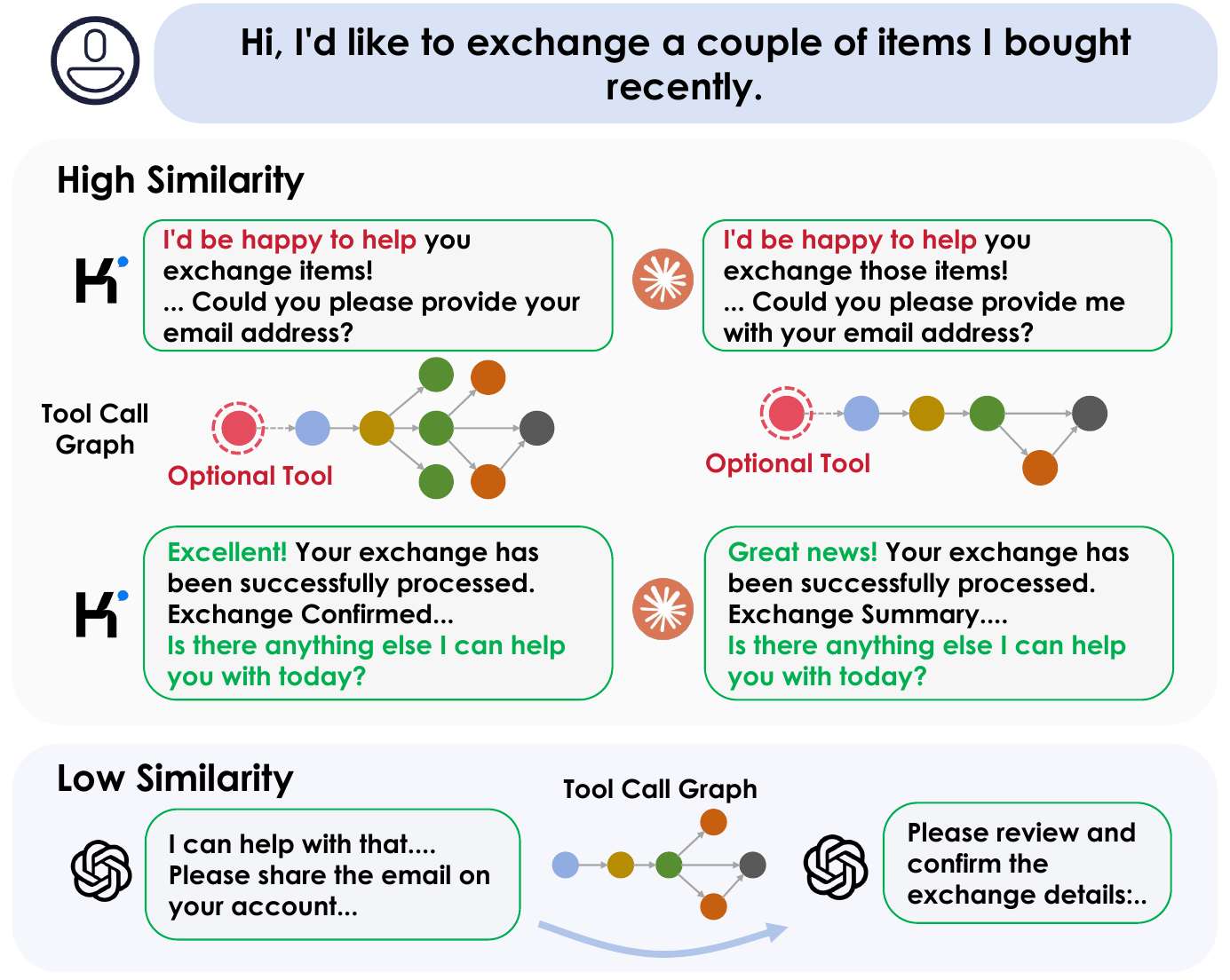}
  \caption{Example trajectories on an item exchange task.}
  \label{fig:intro-a}
\end{subfigure}

\vspace{6pt}

\begin{subfigure}[b]{0.9\columnwidth}
  \centering
  \includegraphics[width=\columnwidth]{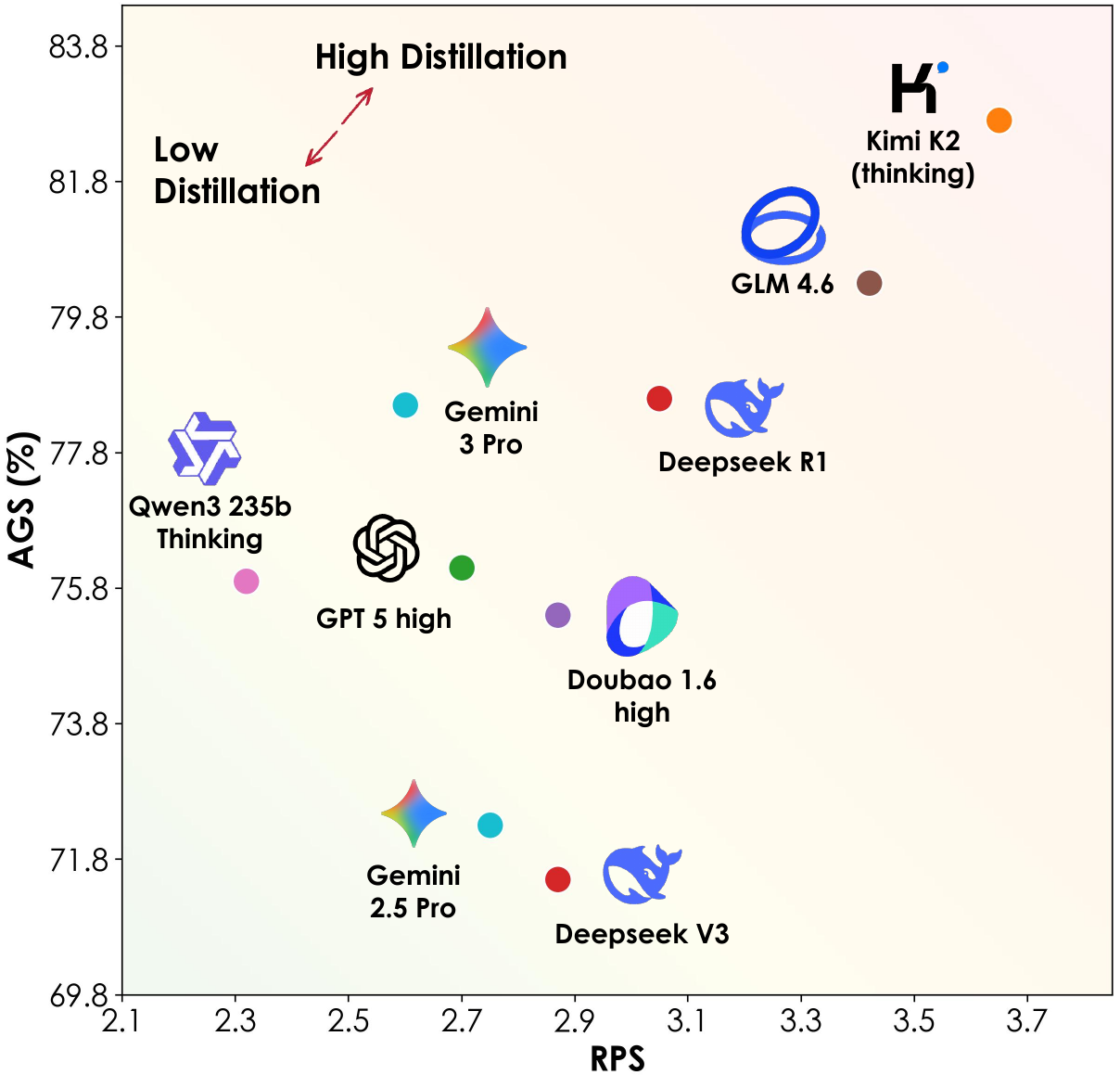}
  \caption{RPS vs.\ AGS across non-Anthropic models.}
  \label{fig:intro-b}
\end{subfigure}

\caption{\textbf{Behavioral similarity to Claude Sonnet 4.5
(thinking).} (a) shows example trajectories from Kimi-K2,
Claude, and GPT-5 on an item exchange task. Kimi-K2 and
Claude share similar response phrasing and optional tool
choices, while GPT-5 adopts a different style. (b) plots
RPS against AGS for all non-Anthropic models. Kimi-K2
(thinking) occupies the top-right corner, closest to the
reference model.}
\label{fig:intro}
\end{figure}

\section{Introduction}\label{sec:introduction}

The current "Cambrian explosion" of high-performing LLM agents often carries a persistent sense of déjà vu.
Despite their diverse origins, many emerging agents exhibit a considerably aligned behavior: they share nearly identical reasoning steps, redundant tool-calling habits, and even the same failure modes~\citep{train,error,Distillingtool}.
This suggests that instead of independent breakthroughs, these models may be distilled echoes of a few dominant teachers. Such pervasive mimicry leads to a convergence of "bad habits" across theoretically independent models~\citep{21,22}. For instance, rather than optimizing for efficiency, many agents mirror the teacher model's verbose reasoning and redundant tool-calling patterns, such as trial-and-error with every available tool, even when the solution is obvious~\citep{31,32}. This collective alignment means that the ecosystem lacks actual robustness; different models no longer provide independent verification but instead fail in the exact same way~\citep{41,42}.

Quantifying this distillation-induced alignment is essential for ensuring ecosystem transparency, but existing methods fall short in the complex, multi-step agent landscape. Current metrics primarily focus on response-level similarity in static dialogues, which fails to capture the dynamic nature of tool-use trajectories~\citep{Quantification2024}.
More critically, they struggle to distinguish between mandatory behaviors related to actions strictly required for task success and non-mandatory behaviors, which reflect a model's autonomous preferences. Without isolating these "behavioral degrees of freedom," it is impossible to determine whether two models converge because there is only one correct path, or because one is blindly shadowing the other.

To bridge this gap, we propose a systematic framework to quantify agent distillation by isolating non-mandatory behavioral patterns.
Our approach introduces two complementary metrics: \textbf{Response Pattern Similarity} (RPS) and \textbf{Action Graph Similarity} (AGS).
RPS measures verbal similarity by segmenting trajectories into five canonical stages (authentication, elicitation, execution, verification, notification) and scoring similarity along style, structure, and alignment dimensions.
AGS measures action-level similarity through three sub-metrics.
$S_{\text{node}}$ captures \textit{optional tool agreement}: for a flight cancellation task, all models must call \texttt{cancel\_reservation}, but some additionally call \texttt{get\_reservation\_details} to double-check.
When two models share such optional choices, they likely share training signals.
$S_{\text{seq}}$ captures \textit{sequential habits} such as post-write verification, pre-modification confirmation, and error retry patterns.
$S_{\text{dep}}$ captures \textit{dependency patterns} such as output reuse rate and tool-chaining depth.
By isolating these non-mandatory behaviors from task-dictated actions, our metrics reveal stylistic and structural alignment that would otherwise be masked by shared correctness.

We evaluate 18 models from 8 major providers on $\tau$-Bench and $\tau^2$-Bench, using Claude Sonnet 4.5 (thinking) as the reference "oracle". Our experiments reveal several key findings.
(1) Anthropic models exhibit strong internal consistency with RPS scores above 3.8, consistent with shared training pipelines.
(2) Kimi-K2 (thinking) shows elevated similarity on both metrics, achieving the highest AGS at 82.7\% among non-Anthropic models, with $S_{\text{node}}$ at 82.6\% and $S_{\text{dep}}$ at 94.7\%.
(3) RPS and AGS capture distinct behavioral dimensions, providing complementary signals for distillation.
(4) A controlled distillation experiment shows that AGS rises toward the teacher and drops toward a non-teacher control, while baseline metrics rise toward both.

Our findings provide empirical grounding for the initial déjà vu. While some models maintain high behavioral diversity, others exhibit elevated similarity with the reference model across both verbal and structural dimensions, even in scenarios where multiple alternative paths exist.

Our main contributions are summarized as follows.
\begin{itemize}[noitemsep, topsep=0pt]
    \item We propose the first framework for tool-use agent distillation quantification, disentangling mandatory and non-mandatory behaviors in tool-use trajectories.
    \item We introduce RPS and AGS, providing interpretable metrics for verbal and action-level behavioral similarity.
    \item We evaluate 18 models from 8 providers, analyzing behavioral convergence patterns in the current agent ecosystem.
\end{itemize}

\section{Related Work}
\noindent
\textbf{Knowledge Distillation for LLM.}
Knowledge distillation transfers knowledge from a large teacher model to a smaller student model~\cite{Distill} and has been widely applied to compress pre-trained models, such as BERT~\cite{distilbert}. For large language models, it is typically categorized into white-box~\cite{PromptDK} and black-box distillation~\cite{alpaca, vicuna2023}. White-box methods require access to intermediate layers or logits~\cite{tinybert, minilm}, whereas black-box approaches use teacher-generated sequences~\cite{GKD, wildchat, Distillsbs}, allowing for distillation from proprietary APIs or arbitrary architectures. However, empirical studies show that distillation can cause homogenization problems in LLMs~\cite{Quantification2024}.

\noindent
\textbf{Data Contamination.}
Data contamination, or benchmark leakage~\cite{Magar022DC, BBLLLM24}, refers to unintentional inclusion of test and benchmark data in training corpora, posing challenges to reliable LLM evaluation~\cite{SainzCGELA23}. Such overlaps inflate performance and undermine fair comparisons~\cite{Magar022DC}. Solutions include contamination detection~\cite{GolchinS25, dekoninck24} and dynamic evaluation~\cite{yu2024freeeval0}. Recent methods include distributional memorization~\cite{dmemwang25} using task-gram language models on semantically related n-grams, and kernel divergence score from embedding kernel similarity matrices~\cite{choi-QDL}.

\noindent
\textbf{Tool-use Benchmark.}
Tool use is a key capability of LLM agents, spurring multiple benchmarks for evaluation. Early benchmarks like API-Bank~\cite{apibank} and Gorilla~\cite{Gorilla} target tool selection and real API calling in large-scale tool sets. ToolBench~\cite{Toolbench24} extends this to multi-step interactions. BFCL~\cite{bfcl} focuses on cross-domain function-calling accuracy. TRAJECT-Bench~\cite{trajbench} introduces trajectory-aware metrics evaluating selection, parameterization, ordering, and dependencies. Domain-specific benchmarks have also emerged, including finance~\cite{finserben} and medicine~\cite{medab}. In conversational agent settings, $\tau$-Bench~\cite{tau-bench} and $\tau^2$-Bench~\cite{tau2-bench} simulate the collaborative use of agent-user tools, emphasizing realistic interactive evaluation.

\begin{figure*}[t]
    \centering
    \includegraphics[width=\textwidth]{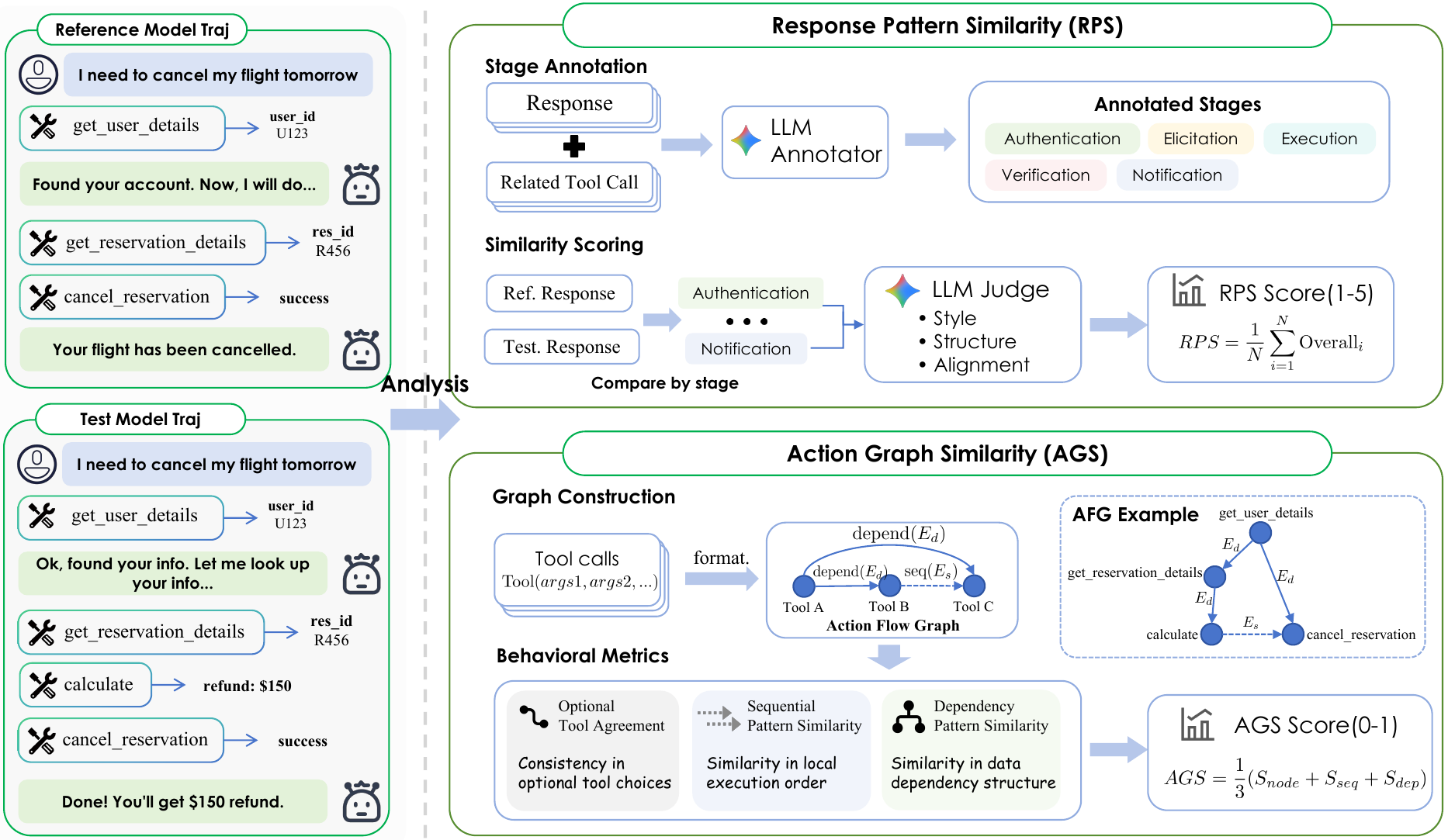}
    \caption{\textbf{Overview of our distillation quantification framework.} \textbf{Left}: Example trajectories from a reference model and a test model on a flight cancellation task. \textbf{Right top}: Response Pattern Similarity (RPS) pipeline, which segments trajectories into canonical stages and scores verbal similarity via an LLM judge. \textbf{Right bottom}: Action Graph Similarity (AGS) pipeline, which constructs Action Flow Graphs from tool call sequences and computes behavioral metrics based on node sets, sequential edges, and dependency edges.}
    \label{fig:method}
\end{figure*}
\section{Method}
\label{sec:method}

\subsection{Trajectory Representation}
\label{subsec:overview}

Given a set of models $\mathcal{M} = \{M_1, ..., M_k\}$ and a tool-use task set $\mathcal{T}$, we collect execution trajectories $\{\tau_1, \tau_2, ...\}$ for each model on $\mathcal{T}$. Each trajectory $\tau$ consists of multiple dialogue turns. Each turn comprises a user input and a model output, where the model output includes (i) user-visible replies, (ii) tool calls, and (iii) tool-related messages such as intermediate outputs or execution traces. Figure~\ref{fig:method} presents the overall framework.

Based on trajectory representation, we design two complementary metrics targeting two distinct information streams within a trajectory. Response Pattern Similarity (RPS) captures verbal fingerprints in how models phrase responses, structure explanations, and express reasoning. Action Graph Similarity (AGS) captures behavioral fingerprints in how models select tools, sequence operations, and reuse outputs. The two metrics capture different aspects of model behavior: models may share verbal patterns but differ in tool usage, or vice versa.

\subsection{Response Pattern Similarity}
\label{subsec:rps}


Response Pattern Similarity (RPS) quantifies verbal similarity between two models along three dimensions. \textit{Style} measures wording habits and vocabulary preferences. \textit{Structure} measures sentence patterns and response templates. \textit{Alignment} measures whether both models exhibit similar reasoning-to-action patterns. Detailed definitions are provided in Appendix~\ref{app:rps}.


Tool-use agent trajectories often span dozens of turns, and different models complete the same task in varying numbers of turns. Directly comparing full trajectories or aligning them turn by turn would match unrelated content, leading to unreliable scores. We therefore design a two-stage pipeline as illustrated in Figure~\ref{fig:method}.


\textbf{Stage Annotation.} To achieve semantic-level alignment across trajectories of different lengths, we define five canonical stages that characterize agent-user interactions: (i) \textit{Authentication} involves identifying the user, verifying identity, or handling a retry after verification failure. (ii) \textit{Elicitation} requests missing parameters from the user to proceed with the task, outside of authentication contexts. (iii) \textit{Execution} invokes domain-specific tools to perform query or modification operations. (iv) \textit{Verification} seeks explicit user confirmation before executing critical write operations. (v) \textit{Notification} reports tool execution results or current status to the user. These five stages are derived from analyzing common interaction patterns in tool-use benchmarks and cover the typical agent workflow. Given a trajectory $\tau$, we use an LLM to annotate each response and tool-related message to its corresponding stage. The annotation prompt and stage definitions are provided in Appendix~\ref{app:annotation}. The scoring procedure is independent of this specific taxonomy; adapting RPS to a new domain requires only replacing the stage definitions.


\textbf{Similarity Scoring.} After stage annotation, RPS compares only the stages shared by both models. Stages appearing in only one trajectory are excluded, as our goal is to measure similarity rather than coverage. For each shared stage, we input the responses, tool-related messages, and associated tool call contents into an LLM judge. The judge scores similarity along the Style, Structure, and Alignment dimensions, as well as a holistic Overall score. Scores range from 1 to 5, where 5 indicates high similarity, 3 indicates moderate overlap, and 1 indicates no discernible similarity. The final RPS score is the mean of the Overall scores across all shared stages. The scoring prompt and detailed rubric are provided in Appendix~\ref{app:scoring}.

\subsection{Action Graph Similarity}
\label{subsec:ags}


Action Graph Similarity (AGS) analyzes structural patterns in tool call sequences to characterize behavioral similarity at the level of tool invocation. Unlike RPS, AGS operates directly on tool call sequences without requiring stage annotation, making it applicable to any domain where agents interact with tools.


\textbf{Graph Construction.} Given a dialogue trajectory $\tau$, we construct a directed graph $G = (V, E_s, E_d)$ based on the tool call sequence.


\textbf{Definition 1 (Tool Node).} Each node $v \in V$ represents a tool call, with attributes:
$$v = (\text{name}, \text{args}, \text{result})$$
where $\text{name}$ is the tool name, $\text{args}$ is the input arguments, and $\text{result}$ is the return value.


\textbf{Definition 2 (Sequential Edge).} A sequential edge $(u, v) \in E_s$ connects temporally adjacent tool calls according to their execution order in $\tau$.


\textbf{Definition 3 (Dependency Edge).} A dependency edge $(u, v) \in E_d$ is established when an argument value of $v$ is derived from the result of $u$. Since simple string matching produces excessive false positives (e.g., common identifiers or dates appearing coincidentally), we use an LLM judge to verify semantic validity. For each candidate edge identified by string overlap, the LLM determines whether the matched value is actually obtained from the source tool's result or is known beforehand (e.g., from user input). We validate the accuracy of this judge on manually annotated edges in Appendix~\ref{app:dep-precision}. When a dependency edge exists between consecutive nodes, the sequential edge is omitted to avoid redundant encoding.

\textbf{Behavioral Metrics.}
Based on the constructed graph $G$, AGS characterizes behavioral similarity along three dimensions. For sequential and dependency patterns, we design three heuristic features, each based on its interpretability.


\textbf{Optional Tool Agreement} $S_{\text{node}}$ measures the consistency of optional tool choices between two models. When completing the same task, models invoke both mandatory tools required for task completion and optional tools for auxiliary purposes. Mandatory tools are dictated by the correct execution path, so all successful models necessarily invoke them. Including these shared invocations in similarity computation inflates scores and obscures model-specific tool preferences. To distinguish these two categories, we analyze successful trajectories from multiple models on the same task. Let $\text{Tools}(M, t)$ denote the set of tools invoked by model $M$ on task $t$, and $\mathcal{M}_t^*$ denote the set of models that successfully complete task $t$. Mandatory tools are the intersection of tool sets across all successful models:
$$\mathcal{F}_t^{\text{mandatory}} = \bigcap_{M \in \mathcal{M}_t^*} \text{Tools}(M, t)$$
Optional tools are those appearing in some but not all successful trajectories. $S_{\text{node}}$ computes the agreement rate on optional tools across all tasks, where agreement means both models either use or skip the same optional tool.


\textbf{Sequential Pattern Similarity} $S_{\text{seq}}$ measures whether two models exhibit similar local execution habits between adjacent tool calls. We extract three features based on sequential edges $E_s$: (i) post-write verification rate, measuring the tendency to invoke a read operation after a write for verification; (ii) pre-write confirmation rate, measuring the tendency to query before executing a write; (iii) error retry rate, measuring the tendency to retry the same tool after an error. Each trajectory is represented as a three-dimensional feature vector. For each task, we compute the cosine similarity between the two models' feature vectors, then average across all tasks to obtain $S_{\text{seq}}$.


\textbf{Dependency Pattern Similarity} $S_{\text{dep}}$ measures whether two models exhibit similar patterns in reusing tool outputs. We extract three features based on dependency edges $E_d$: (i) output reuse rate, measuring the proportion of tool calls that reuse outputs from preceding calls; (ii) longest dependency chain length, measuring the depth of chained planning; (iii) output fan-out rate, measuring the proportion of tool outputs reused by multiple subsequent calls. Similar to $S_{\text{seq}}$, we compute per-task cosine similarity and average across tasks.
Detailed definitions and computation methods for the three sub-metrics are provided in Appendix~\ref{app:ags}.

\begin{table*}[ht]
\centering

\caption{Behavioral similarity of 18 models to Claude Sonnet 4.5 (thinking) across four baseline metrics, three AGS sub-metrics, and three RPS dimensions. Anthropic models (shaded) serve as within-family baselines. Among non-Anthropic models: \textbf{bold} = 1st, \textit{italic} = 2nd, \underline{underline} = 3rd. Sty.=Style, Str.=Structure, Ali.=Alignment.}
\small
\label{tab:trace-similarity}
\setlength{\tabcolsep}{2.5pt}
\renewcommand{\arraystretch}{1.15}
\begin{tabular}{@{}l l cccc c cccc c cccc@{}}
\toprule
\multirow{2}{*}{\textbf{Family}} & \multirow{2}{*}{\textbf{Model}} & \multicolumn{4}{c}{\textbf{Baseline}} & & \multicolumn{4}{c}{\textbf{AGS (\%)}} & & \multicolumn{4}{c}{\textbf{RPS}} \\
\cmidrule(lr){3-6} \cmidrule(lr){8-11} \cmidrule(lr){13-16}
& & GED & RSE & N-gram & BERT & & $S_{\text{node}}$ & $S_{\text{seq}}$ & $S_{\text{dep}}$ & \textbf{Avg} & & Sty. & Str. & Ali. & \textbf{Overall} \\
\midrule

\rowcolor{colorAnthropic}
\textsc{Anthropic} & Sonnet 4.5 (no-think) & 82.6 & 3.14 & 0.371 & 0.951 & & 79.1 & 79.9 & 94.0 & 84.3 & & 4.07 & 3.89 & 3.66 & 3.87 \\
\rowcolor{colorAnthropic}
 & Opus 4.1 (thinking) & 79.7 & 3.25 & 0.355 & 0.946 & & 81.0 & 74.4 & 93.7 & 83.0 & & 4.02 & 3.80 & 3.72 & 3.85 \\
\midrule

\rowcolor{colorOpenAI}
\textsc{OpenAI} & GPT-4.1 & 75.2 & 2.45 & 0.306 & 0.936 & & \underline{75.9} & \textit{74.6} & 88.0 & \underline{79.5} & & 3.07 & 2.92 & 3.45 & 3.15 \\
\rowcolor{colorOpenAI}
 & GPT-5 & 68.3 & 2.26 & 0.230 & 0.889 & & 71.3 & 69.4 & 87.7 & 76.1 & & 2.50 & 2.43 & 3.17 & 2.70 \\
\midrule

\rowcolor{colorDeepSeek}
\textsc{DeepSeek} & R1 & 70.8 & 2.44 & 0.259 & 0.925 & & \textit{78.3} & \underline{72.5} & 85.0 & 78.6 & & 2.99 & 2.87 & 3.29 & 3.05 \\
\rowcolor{colorDeepSeek}
 & V3.1 (thinking) & 72.6 & 2.56 & \underline{0.307} & \underline{0.932} & & 78.1 & 63.5 & \underline{91.2} & 77.6 & & \underline{3.24} & 2.91 & \textit{3.43} & 3.19 \\
\rowcolor{colorDeepSeek}
 & V3.1 (no-think) & 69.2 & 2.29 & \textit{0.312} & \textit{0.931} & & 72.4 & 65.3 & 87.5 & 75.1 & & \textit{3.55} & \textit{3.27} & 3.38 & \underline{3.40} \\
\rowcolor{colorDeepSeek}
 & V3-0324 & 59.1 & 1.87 & 0.270 & 0.917 & & 77.5 & 62.7 & 74.5 & 71.5 & & 2.84 & 2.66 & 3.13 & 2.87 \\
\midrule

\rowcolor{colorMoonshot}
\textsc{Moonshot} & Kimi-K2 (thinking) & \textit{78.1} & \textit{2.81} & \textbf{0.343} & \textbf{0.938} & & \textbf{82.6} & 70.8 & \textbf{94.7} & \textbf{82.7} & & \textbf{3.86} & \textbf{3.57} & \underline{3.51} & \textbf{3.65} \\
\rowcolor{colorMoonshot}
 & Kimi-K2 & \underline{74.8} & \underline{2.79} & 0.318 & 0.925 & & 70.2 & \textbf{73.3} & \textit{90.4} & 77.9 & & 3.67 & \underline{3.33} & 3.38 & \textit{3.46} \\
\midrule

\rowcolor{colorByteDance}
\textsc{ByteDance} & Doubao 1.6 (high) & 70.3 & 2.32 & 0.287 & 0.930 & & 68.5 & 69.5 & 88.1 & 75.4 & & 2.69 & 2.57 & 3.36 & 2.87 \\
\rowcolor{colorByteDance}
 & Doubao 1.6 (medium) & 72.9 & 2.24 & 0.276 & 0.931 & & 76.9 & 67.3 & 88.3 & 77.5 & & 2.71 & 2.56 & 3.38 & 2.88 \\
\rowcolor{colorByteDance}
 & Doubao 1.6 (low) & 71.8 & 2.07 & 0.274 & 0.928 & & 76.7 & 68.0 & 88.8 & 77.8 & & 2.59 & 2.47 & 3.33 & 2.80 \\
\midrule

\rowcolor{colorGoogle}
\textsc{Google} & Gemini 3 Pro & \textbf{77.5} & 2.20 & 0.294 & 0.924 & & 71.9 & 71.2 & 92.3 & 78.5 & & 2.68 & 2.31 & 2.81 & 2.60 \\
\rowcolor{colorGoogle}
 & Gemini 2.5 Pro & 70.1 & 2.24 & 0.281 & 0.922 & & 71.7 & 64.6 & 80.5 & 72.3 & & 2.65 & 2.42 & 3.18 & 2.75 \\
\midrule

\rowcolor{colorQwen}
\textsc{Qwen} & Qwen3-30B (thinking) & 65.8 & 2.51 & 0.250 & 0.895 & & 77.9 & 62.4 & 89.9 & 76.7 & & 2.65 & 2.42 & 2.76 & 2.42 \\
\rowcolor{colorQwen}
 & Qwen3-235B (thinking) & 65.6 & 2.05 & 0.245 & 0.897 & & 68.1 & 67.2 & 92.4 & 75.9 & & 2.62 & 2.43 & 2.77 & 2.40 \\

\midrule

\rowcolor{colorZhipu}
\textsc{Zhipu} & GLM-4.6 & 75.9 & \textbf{2.71} & 0.306 & 0.925 & & 80.4 & 71.9 & 88.7 & \textit{80.3} & & 3.45 & 3.26 & \textbf{3.54} & 3.42 \\

\bottomrule
\end{tabular}
\end{table*}

\section{Experiments}
\label{sec:experiments}

\subsection{Experiment Settings}
\label{subsec:settings}

\textbf{Model Families.} We evaluate 18 models from 8 major providers: Anthropic (Claude Sonnet 4.5~\citep{anthropic_claude45sonnet}, Claude Opus 4.1~\citep{anthropic_claude41opus}), OpenAI (GPT-4.1~\citep{openai_gpt41}, GPT-5~\citep{openai_gpt5}), DeepSeek (R1~\citep{deepseek_r1}, V3.1~\citep{deepseek_v3_1226}, V3-0324~\citep{deepseek_v3_0324}), Moonshot (Kimi-K2~\citep{Moonshot_kimik2}), ByteDance (Doubao 1.6~\citep{bytedance_doubao16}), Google (Gemini 2.5 Pro~\citep{google_gemini25pro0506}, Gemini 3 Pro~\citep{google_gemini3pro}), Qwen (Qwen3-30B~\citep{qwen3_30b}, Qwen3-235B~\citep{qwen3_235b}), and Zhipu (GLM-4.6~\citep{zhipu_glm4_6}). All models are accessed via official APIs with default hyperparameters. We verify that temperature-induced AGS variation across runs remains substantially smaller than cross-model differences (Appendix~\ref{app:temp}). We use Claude Sonnet 4.5 (thinking) as the reference model and compute behavioral similarity between each model and the reference.


\textbf{Tool-use Benchmark.} We use $\tau$-Bench~\citep{tau-bench} and $\tau^2$-Bench~\citep{tau2-bench} as the evaluation benchmark, which contains real-world agent tasks across three domains: airline and retail from $\tau$-Bench, and telecom from $\tau^2$-Bench. We sample 50 tasks from each domain, covering typical scenarios such as identity verification, information retrieval, and order modification.

\textbf{RPS Evaluation.} For RPS annotation and scoring, we use Gemini-2.5-flash-thinking as the LLM annotator and judge with default temperature. 
Model-level rankings remain stable across temperature settings (Appendix~\ref{app:judge-temp}).


\textbf{Baseline metrics.} We compare against two categories of baselines.
Semantic baselines include (i) \textbf{RSE}~\citep{Quantification2024}, computed on model responses following prior work,
(ii) \textbf{$n$-gram overlap} with $n{=}2$ (2-gram overlap ratio),
and (iii) \textbf{BERTScore}~\citep{bertscore}, a contextual-embedding-based semantic similarity.
For graph structure, we use (iv) \textbf{GED}~\citep{GED}, with similarity $1 - d_{\mathrm{GED}}/(|V_1|+|V_2|+|E_1|+|E_2|)$.
\begin{figure*}[t]
    \centering
    \includegraphics[width=\textwidth]{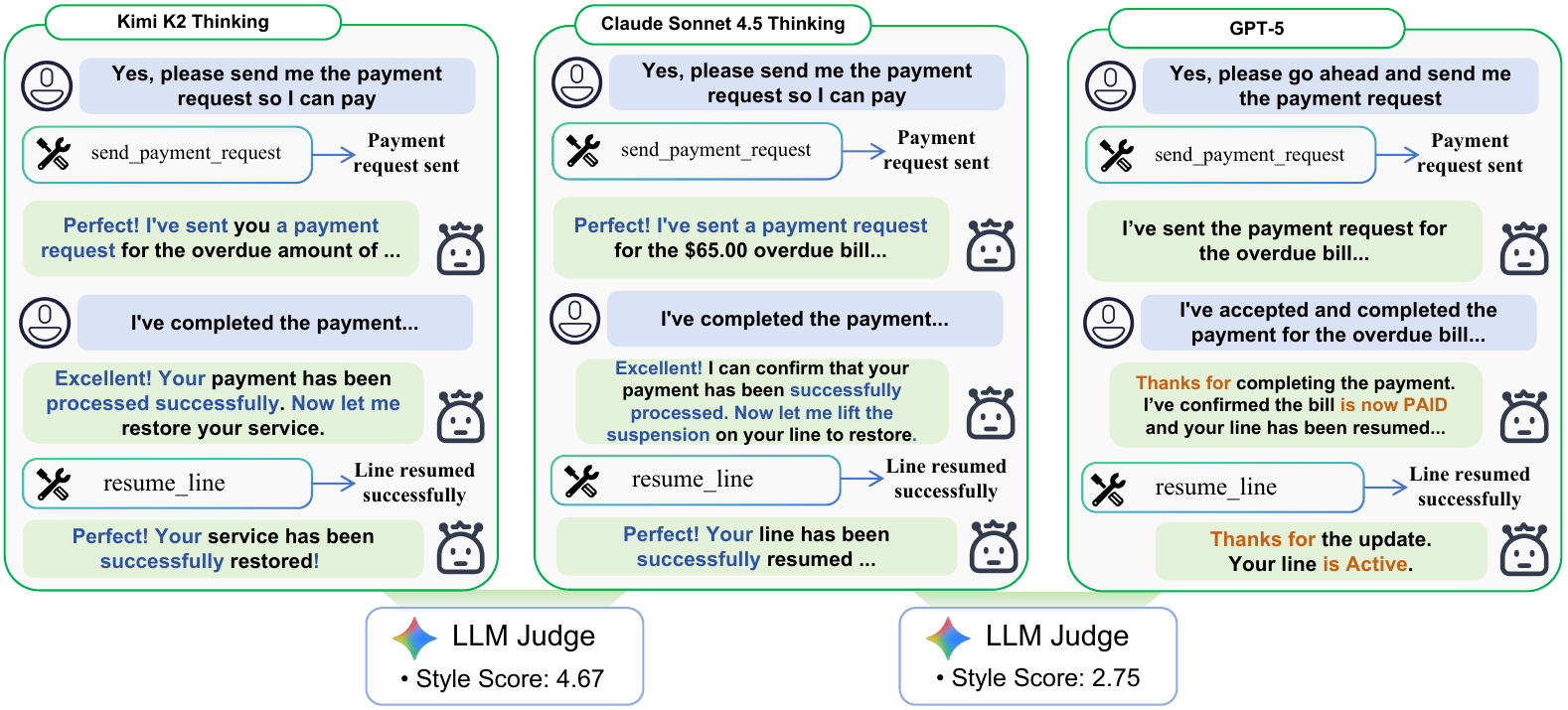}
    \caption{RPS case study on a telecom customer service task. \textbf{Left.} Kimi-K2 (thinking) trajectory. \textbf{Middle.} Claude Sonnet 4.5 (thinking) as reference. \textbf{Right.} GPT-5 trajectory. Style scores from the LLM judge are shown at the bottom of each panel.}
    \label{fig:case-study-style}
\end{figure*}
\subsection{Main Results}
\label{subsec:results}


Table~\ref{tab:trace-similarity} presents the behavioral similarity between each model and Claude Sonnet 4.5 (thinking). AGS measures action-level patterns, while RPS measures verbal patterns.

\textbf{Anthropic models exhibit strong internal consistency.} Both Anthropic models achieve RPS scores above 3.8, with Sonnet 4.5 (no-think) at 3.87 and Opus 4.1 (thinking) at 3.85. These scores exceed those of all non-Anthropic models by at least 0.20 points. The two models also achieve $S_{\text{dep}}$ scores of 94.0\% and 93.7\% respectively, indicating highly similar tool invocation preferences. This within-family consistency aligns with expectations for models sharing training pipelines and serves as a baseline for interpreting cross-family similarity (Section~\ref{subsec:lineage}).

\textbf{Kimi-K2 (thinking) exhibits exceptionally high similarity to the reference model.} Among non-Anthropic models, Kimi-K2 (thinking) achieves the highest AGS at 82.7\% and the highest RPS at 3.65. Notably, its $S_{\text{node}}$ reaches 82.6\% and $S_{\text{dep}}$ reaches 94.7\%, both exceeding the Anthropic baseline (Opus 4.1 at 81.0\% and 93.7\%). This means Kimi-K2 (thinking) shares more optional tool choices and tool invocation preferences with the reference model than models from the same provider family. When we repeat the analysis using GPT-5 as an alternative reference, Kimi-K2's similarity to Claude remains consistently higher than its similarity to GPT-5, confirming that this pattern reflects directional behavioral inheritance rather than benchmark-induced convergence across all models (Appendix~\ref{app:multi-ref}).

\textbf{Sub-metrics capture different behavioral dimensions.} The three AGS sub-metrics measure distinct aspects of tool-use behavior. $S_{\text{node}}$ measures optional tool selection preferences, where Kimi-K2 (thinking) exhibits the highest similarity to the reference model at 82.6\%. Sensitivity analysis shows $S_{\text{node}}$ remains stable across different model subsets (Appendix~\ref{app:bootstrap}). Without mandatory/optional separation, $S_{\text{node}}$ inflates by 12.2 pp on average for non-family models (Appendix~\ref{app:mandatory-ablation}).
$S_{\text{seq}}$ captures local execution habits such as post-write verification, pre-write confirmation, and error persistence patterns. $S_{\text{dep}}$ measures dependency topology such as output reuse rate, dependency chain depth, and output fan-out rate. Kimi-K2 (thinking) also exhibits the highest $S_{\text{dep}}$ similarity at 94.7\%, suggesting that its tool dependency patterns closely mirror those of the reference model. Stage alignment reduces within-case RPS scoring variance by 44--55\% compared to full-trajectory comparison (Appendix~\ref{app:stage-ablation}). 

\begin{table}[t]
\centering
\caption{AGS and RPS for within-family and cross-family model pairs on 50 randomly sampled tasks. Within-family pairs score 5.9 pp higher in AGS than the cross-family pair.}
\label{tab:lineage}
\small
\resizebox{\columnwidth}{!}{%
\begin{tabular}{@{}llcc@{}}
\toprule
\textbf{Type} & \textbf{Model Pair} & \textbf{AGS (\%)} & \textbf{RPS} \\
\midrule
\multirow{2}{*}{Within-family}
& Opus 4.1 -- Sonnet 4.5 & 87.2 & 4.18 \\
& DeepSeek-R1 -- V3.1 (thinking) & 86.7 & 3.76 \\
\midrule
Cross-family
& DeepSeek-R1 -- Sonnet 4.5 & 81.1 & 3.48 \\
\bottomrule
\end{tabular}%
}
\end{table}

\begin{figure}[t]
\centering
\includegraphics[width=\columnwidth]{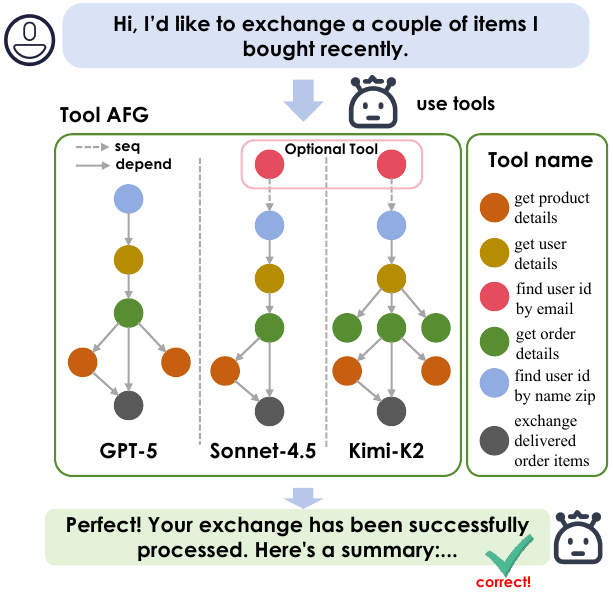}
\caption{\textbf{Action Flow Graph comparison on an item exchange task.} All three models complete the task successfully. Claude Sonnet 4.5 (thinking) and Kimi-K2 first invoke the optional \texttt{find\_user\_id\_by\_email} (red node), whereas GPT-5 skips this step and directly calls \texttt{find\_user\_id\_by\_name\_zip}. Dashed arrows denote sequential edges; solid arrows denote dependency edges.}
\label{fig:ags-case-study}
\end{figure}
\section{Discussion}
\label{sec:discussion}
\subsection{AGS and RPS Detect Behavioral Similarity Patterns}
\label{subsec:lineage}


Models with confirmed distillation relationships, such as the DeepSeek-R1-Distill-Qwen series, are distilled exclusively for reasoning and achieve low success rates on $\tau$-Bench and $\tau^2$-Bench, precluding trajectory-level analysis. To obtain ground-truth validation, we first compare within-family and cross-family model pairs to establish a behavioral baseline, then conduct a controlled distillation experiment with a known teacher-student pair.

We first examine within-family similarity as a baseline. Models from the same provider share training pipelines and are expected to exhibit higher behavioral similarity than cross-family pairs. When a cross-family pair reaches comparable similarity, it suggests behavioral inheritance beyond independent development. We randomly sample 50 tasks from the benchmark for this analysis.

\textbf{Within-family pairs achieve 5.9 pp higher AGS than cross-family pairs.} As shown in Table~\ref{tab:lineage}, Claude Opus 4.1 and Claude Sonnet 4.5 achieve 87.2\% AGS and 4.18 RPS within the Anthropic family. Similarly, DeepSeek-R1 and DeepSeek-V3.1 (thinking) achieve 86.7\% AGS and 3.76 RPS within the DeepSeek family. By contrast, the cross-family pair DeepSeek-R1 versus Claude Sonnet 4.5 achieves only 81.1\% AGS and 3.48 RPS. This 5.9 pp gap confirms that our metrics capture behavioral inheritance, establishing a baseline for interpreting cross-family similarity in the following analysis.

\begin{table}[t]
\centering
\caption{Controlled distillation validation. Qwen2.5-14B-Instruct is fine-tuned on 200 Claude Sonnet 4.5 trajectories via LoRA. Similarity is measured toward the teacher (Claude) and a non-teacher control (DeepSeek R1).}
\label{tab:distill-valid}
\small
\resizebox{\columnwidth}{!}{%
\begin{tabular}{@{}llccc@{}}
\toprule
\textbf{Direction} & \textbf{Metric} & \textbf{Baseline} & \textbf{Distilled} & \textbf{$\Delta$} \\
\midrule
\multirow{2}{*}{$\rightarrow$ Teacher (Claude)} & AGS & 0.59 & 0.72 & +0.13 \\
 & GED & 0.42 & 0.65 & +0.23 \\
\midrule
\multirow{2}{*}{$\rightarrow$ Control (R1)} & AGS & 0.64 & 0.59 & $-$0.05 \\
 & GED & 0.39 & 0.59 & +0.20 \\
\bottomrule
\end{tabular}%
}
\end{table}

\textbf{Controlled distillation produces a directional signal in AGS.} To further validate our metrics under known ground truth, we fine-tune Qwen2.5-14B-Instruct on 200 $\tau$-bench trajectories generated by Claude Sonnet 4.5 via LoRA, with DeepSeek R1 as a non-teacher control (training details in Appendix~\ref{app:distill-setup}). As shown in Table~\ref{tab:distill-valid}, AGS toward the teacher increases by +0.13 while AGS toward the control decreases by $-$0.05. In contrast, GED increases by +0.23 and +0.20 toward the teacher and control respectively, with a gap of only 0.03. Sub-metric decomposition and RPS analysis are provided in Appendix~\ref{app:distill-analysis}.

\begin{table}[t]
\centering
\caption{AGS sub-metrics and GED for three model pairs across task outcome settings (both-correct, both-wrong, mixed). Claude = Claude Sonnet 4.5 (thinking), Kimi = Kimi-K2 (thinking), GPT = GPT-5. Bold marks the highest value per column within each setting.}
\label{tab:ags_submetrics}
\resizebox{\columnwidth}{!}{
\begin{tabular}{llcccc}
\toprule
Setting & Model Pair & GED & $S_{\text{node}}$ & $S_{\text{seq}}$ & $S_{\text{dep}}$ \\
\midrule
\multirow{3}{*}{Both-correct}
& Kimi -- Claude & \textbf{0.814} & \textbf{0.661} & 0.892 & \textbf{0.993} \\
& GPT -- Claude  & 0.705 & 0.196 & 0.904 & 0.879 \\
& Kimi -- GPT    & 0.737 & 0.143 & \textbf{0.912} & 0.882 \\
\midrule
\multirow{3}{*}{Both-wrong}
& Kimi -- Claude & 0.558 & 0.355 & \textbf{0.829} & \textbf{0.953} \\
& GPT -- Claude  & 0.522 & 0.258 & 0.680 & 0.899 \\
& Kimi -- GPT    & \textbf{0.594} & \textbf{0.387} & 0.583 & 0.888 \\
\midrule
\multirow{3}{*}{Mixed}
& Kimi -- Claude & \textbf{0.718} & \textbf{0.414} & 0.572 & \textbf{0.917} \\
& GPT -- Claude  & 0.599 & 0.279 & 0.562 & 0.873 \\
& Kimi -- GPT    & 0.620 & 0.307 & \textbf{0.633} & 0.829 \\
\bottomrule
\end{tabular}
}
\end{table}
\subsection{Kimi-K2 Exhibits Claude-like Behavioral Patterns}
\label{subsec:case-study}


The baseline established above reveals an unexpected pattern. Kimi-K2 (thinking) achieves 82.7\% AGS and 3.65 RPS with Claude Sonnet 4.5 (thinking), the highest among all non-Anthropic models. Notably, this cross-family similarity exceeds some within-family Anthropic pairs, raising the question of whether Kimi-K2 has inherited behavioral patterns from Claude. We investigate this hypothesis through case studies on both response style and tool invocation.


\textbf{Kimi-K2 and Claude share conversational style patterns that GPT-5 does not exhibit.} Figure~\ref{fig:case-study-style} compares response trajectories in a customer service scenario. Unlike GPT-5, which produces brief, procedurally oriented responses, both Claude Sonnet 4.5 (thinking) and Kimi-K2 consistently employ informal phrasing with enthusiastic affirmatives such as "Excellent!" and "Perfect!". Even without explicit directives in the system prompt, both models proactively acknowledge the current state, anticipate subsequent steps, and provide emotional support. This shared tendency toward user engagement, absent in GPT-5, suggests that it is inherited conversational heuristics rather than task-specific adaptation. Additional case studies are provided in Appendix~\ref{app:rps-case}.


\textbf{Kimi-K2 and Claude share optional tool preferences that GPT-5 does not exhibit.} Figure~\ref{fig:ags-case-study} illustrates a concrete case. In an exchange request task, both Claude and Kimi-K2 first invoke the optional \texttt{find\_user\_id\_by\_email}, while GPT-5 skips this step and directly calls \texttt{find\_user\_id\_by\_name\_zip}. Email verification is not required for task completion; however, Claude and Kimi-K2 share a preference for redundant verification. As shown in Table~\ref{tab:ags_submetrics}, the Kimi--Claude pair achieves $S_{\text{node}}$ of 0.661 in the both-correct setting, 3.4$\times$ higher than 0.196 for the GPT--Claude pair. This pattern remains stable across "both-wrong" and mixed settings, suggesting that the similarity reflects inherent decision-making heuristics rather than task-specific strategies. Detailed analysis of $S_{\text{seq}}$ and $S_{\text{dep}}$ is provided in Appendix~\ref{app:ags-case}.


Taken together, Kimi-K2 exhibits high similarity to Claude across both verbal expression (RPS 3.65) and tool invocation (AGS 82.7\%), reaching levels comparable to within-family pairs on several sub-metrics. The two metrics capture largely independent signals (Pearson $r$ = 0.491, $p$ = 0.054 across 16 non-Anthropic models), confirming that this convergence spans distinct behavioral dimensions. Correlation analysis with baseline metrics and additional ablation studies are provided in Appendix~\ref{app:correlation}.

\section{Conclusion}
We present a framework for quantifying behavioral similarity in LLM agents by disentangling mandatory task behaviors from non-mandatory patterns through two complementary metrics, RPS and AGS. Evaluating 18 models from 8 providers, we find that within-family model pairs consistently score higher than cross-family pairs. Among cross-family pairs, Kimi-K2 (thinking) stands out with $S_{\text{node}}$ at 82.6\% and $S_{\text{dep}}$ at 94.7\%, both exceeding Anthropic's own Opus 4.1. A controlled distillation experiment validates the metrics: AGS increases toward the known teacher while decreasing toward a control, whereas GED rises in both directions without distinguishing the two. We release our code to support further work on behavioral diagnostics in the agent ecosystem.

\section*{Acknowledgements}
\label{sec: Acknowledgements}
This work was supported by the National Natural Science Foundation of China (No. 62121002) and the advanced computing resources provided by the Supercomputing Center of the USTC.
\section*{Limitations}

\textbf{Pairwise Analysis Scope.}
Our framework can construct a full pairwise similarity matrix,
but we report results relative to a single reference model due
to computational cost. Complete pairwise analysis across 18
models would require 153 comparisons.

\textbf{Benchmark Coverage.}
We evaluate on $\tau$-Bench and $\tau^2$-Bench, covering three
English-language customer service domains (airline, retail,
telecom). Generalizability to other domains, task types, or
languages remains to be validated.

\textbf{Scope of Applicability.}
RPS and AGS are designed for tool-use agents with observable
tool invocations. AGS operates on tool-call sequences and
applies wherever tools are used. RPS depends on a
domain-specific stage taxonomy; adapting it to domains with
different interaction patterns (e.g., coding agents with
iterative self-reflection) requires new stage definitions.
Extending the framework to non-tool-use paradigms such as code
generation or multi-agent collaboration would require further
methodological work.

\textbf{Validation.}
We validate our metrics through within-family comparison and a
controlled distillation experiment. Validation on publicly
deployed models with confirmed distillation relationships remains
open, as such models currently lack sufficient tool-use
capabilities for trajectory-level analysis. Additionally,
framework choice can affect behavioral similarity; we use a
generic ReAct framework to avoid masking distillation signals
(Appendix~\ref{app:framework}).
\bibliography{custom}

\appendix
\section{Response Pattern Similarity: Pipeline and Prompts}
\label{app:rps}

RPS proceeds in three steps: stage annotation, trajectory alignment, and stage-level similarity scoring.

\subsection{Stage Annotation}
\label{app:annotation}

An LLM assigns each model output to a task stage. The prompt template is shown below. The five stages below cover the canonical interaction structure in the evaluated tool-use benchmarks. The scoring pipeline is agnostic to this particular taxonomy; applying RPS to a different domain requires only replacing the stage definitions and annotation prompt.

\begin{itemize}[noitemsep, topsep=0pt]
    \item \textbf{Authentication.} Identifying the user, verifying identity, or handling a retry after verification failure.
    \item \textbf{Elicitation.} Requesting missing parameters needed to proceed with the task, outside the verification phase.
    \item \textbf{Execution.} Invoking specific business tools to perform query or modification operations.
    \item \textbf{Verification.} Seeking explicit user confirmation before executing critical write operations.
    \item \textbf{Notification.} Reporting tool execution results or current status to the user.
\end{itemize}

\begin{Prompt}[label={prompt:basic},title={System Prompt for Stage Annotation}]

# Role

You are an expert in agent trajectory semantic annotation. Your responsibility is to annotate the semantic stage (intent) of each think/response in an LLM agent's task flow, enabling trajectory alignment across different models at the semantic level.

# Task

Annotate the intent category for a given think/response. When different models complete the same task, the number and order of steps may vary, but semantic stages remain relatively stable. Through intent annotation, we can align trajectories from different models on a logical timeline for cross-model behavior comparison.

# Input Data

## Data Structure
An assistant's single-turn message consists of multiple content items:
- **think**: Internal reasoning (not directly shown to user)
- **response**: Natural language reply to the user
- **tool_call**: Tool invocation (name / arguments / result)

Example:
```
[think 1]: "The user needs to query their account, I should verify their identity first"
[tool_call 1]: {"name": "user_lookup", "arguments": "...", "result": "..."}
[response 1]: "Let me confirm your email address"
```

# Intent Categories

Based on the core purpose of the think/response in task progression, annotate as one of the following five categories:

1. **Authentication**: Identity verification. Identifying the user, verifying identity, or handling retry after verification failure.
2. **Elicitation**: Information gathering. In non-authentication contexts, requesting missing parameters from the user needed to proceed with the task.
3. **Execution**: Task execution. Invoking domain-specific tools to perform query or modification operations.
4. **Verification**: Operation confirmation. Seeking explicit user confirmation before executing critical write operations.
5. **Notification**: Result reporting. Reporting tool execution results or current status to the user.

# Annotation Principles

1. **Primary intent takes precedence**: If multiple behaviors are present (e.g., explaining + asking), prioritize the main purpose.
2. **Authentication takes precedence**: As long as the process of establishing user identity is ongoing, it must be annotated as Authentication.
3. **Fact-based**: The reason field must be based on observable content, without speculation.

# Output Format

```json
{
    "reason": "Brief explanation of classification rationale",
    "intent": "Authentication|Elicitation|Execution|Verification|Notification"
}
```
\end{Prompt}

\subsection{Trajectory Alignment}
\label{app:alignment}

For each trajectory, all \texttt{think}/\texttt{response} items are
grouped by their annotated intent label. Items within each group
are augmented with adjacent tool-call context and concatenated
into a single block. For each intent label shared by both
trajectories, the two blocks are sent together to the LLM judge
as one scoring unit. Labels appearing in only one trajectory are
excluded.

\subsection{Similarity Evaluation}
\label{app:scoring}

Each shared stage is then scored by an LLM judge along Style, Structure, and Alignment. The evaluation prompt is shown below.

\begin{Prompt}{System Prompt for LLM Judge}
# Role

You are an expert in agent behavior alignment analysis, specializing in detecting behavioral fingerprints of model distillation. Your task is to compare the expression patterns of two models at the same task stage to identify potential distillation signals.

# Task

Compare the think/response similarity between Reference Model (Model A) and the target Model (Model B) at the same intent stage. Focus on **spontaneous convergence not required by instructions** to determine whether Model B exhibits behavioral fingerprints of being distilled from Model A.

# Input Data

## Data Structure
The input contains all think/response items from both models at the same intent stage. Each item includes:
- **type**: think (internal reasoning) or response (reply to user)
- **content**: natural language content
- **context**: tool_call before/after this think/response (if any)

## Analysis Framework
Before scoring, analyze the behavioral patterns of both models along the following dimensions:

### Reasoning Pattern
- **Constraint_Check**: Explicitly references and checks policies or rules from the System Prompt
- **Decomposition**: Breaks down complex requests into multiple sub-steps
- **Error_Recovery**: Self-correction after encountering API errors or information mismatch
- **Linear_Forward**: Standard reasoning process without explicit rule references, decomposition, or error correction

### Alignment Pattern
Assess the consistency between the tool-calling intent expressed in think/response and the actual execution.

**Judgment Logic**: First determine whether an explicit tool-calling intent is expressed (e.g., "I will query...", "Let me call...").

- **Consistent**: Intent matches execution, or no intent and no execution
- **Tool_Mismatch**: Declared to call tool A, but actually called tool B
- **Param_Mismatch**: Correct tool, but parameter values differ from declaration
- **Omission**: Declared to call a tool, but did not execute subsequently
- **Hallucinated_Action**: No declared intent, but tool_call was executed

**Important Judgment Rules**:
1. **Consecutive thinking is not Omission**: If think1 declares intent, think2 continues reasoning, then tool_call executes, this is a normal multi-step thinking process, not Omission. Only when the declared tool call is never executed in the entire turn should it be considered Omission.
2. **Historical context is not Hallucinated_Action**: If the tool_call's parameters or purpose can be traced back to information or intent from previous conversation turns, this is not hallucinated execution. Hallucinated_Action only refers to tool calls that appear out of nowhere within the current turn and cannot be explained by context.
3. **Delayed execution is normal**: In structures like think1-think2-toolcall1-toolcall2, it is normal for think2's idea to be executed in toolcall2, and should not be judged as anomalous.

### Structure Pattern
- **Template_Based**: Follows strict, repetitive sentence patterns, such as fixed scripts
- **Free_Form**: No obvious fixed templates, flexible and varied sentence structures

# Scoring Dimensions

## 1. Style (Wording Style Similarity)
The degree of similarity in expression methods and vocabulary habits spontaneously adopted by both models.

**Focus on**:
- Wording of opening and closing statements
- Information organization (what to say first and last)
- Common vocabulary and phrases
- Tone characteristics (formal/colloquial)

**Distillation Signal**: Distilled models inherit the vocabulary habits of the teacher model.

## 2. Structure (Sentence Structure Similarity)
The degree of similarity in syntactic structures and response templates between the two models.

**Focus on**:
- Paragraph organization (lists vs paragraphs, numbering style)
- Information granularity (level of detail, which fields to display)
- Reasoning presentation (how to present Decomposition or Constraint_Check)
- Template usage (whether both use Template_Based or both use Free_Form)

**Distillation Signal**: Distilled models inherit the expression templates of the teacher model.

## 3. Alignment (Execution Alignment Similarity)
Whether the two models exhibit similar behavioral patterns in think-execution alignment.

**Focus on**:
- Whether alignment pattern types are the same (both Consistent, or both have a specific anomaly)
- Think-to-execution transition style (e.g., both batch-execute after multi-step thinking, or both execute step-by-step)
- Timing and decision patterns of tool calls
- Whether error recovery strategies are similar

**Scoring Logic**:
- If both models are Consistent with similar think-execution style -> high score
- If both models have **the same type of anomaly** -> high score (strongest distillation signal)
- If one is normal and one is anomalous, or anomaly types differ -> low score

**Distillation Signal**: Distilled models inherit the execution habits and decision patterns of the teacher model.

# Scoring Principles

1. **Exclude instruction requirements**: Formats/wording required by System Prompt or few-shot examples should not count toward similarity
2. **Same anomaly is a strong signal**: If both models exhibit the same type of execution anomaly, this is the strongest distillation signal
3. **Non-standard convergence is more valuable**: Both models adopting the same non-standard expression is more indicative of similarity than both adopting standard approaches
4. **Holistic judgment**: High similarity in any dimension can justify a high overall score
5. **Cautious anomaly judgment**: Before determining Omission or Hallucinated_Action, verify whether exceptions in the "Important Judgment Rules" apply

# Scoring Rubric

- **5**: Very similar. Highly consistent wording style and sentence structure, or identical anomaly patterns
- **4**: Similar. Multiple similarities but with some differences
- **3**: Neutral. Some similarities but insufficient to determine as similar
- **2**: Dissimilar. No significant similarities, with obvious differences
- **1**: Very dissimilar. Completely different wording styles and expression methods

# Output Format

```json
{
    "analysis": {
        "style": "Compare the wording styles of both models, pointing out specific similarities or differences",
        "structure": "Compare the sentence structures of both models, pointing out specific similarities or differences",
        "alignment": "Analyze the execution alignment patterns of both models, label each model's Alignment Pattern type, and explain whether think-execution styles are similar"
    },
    "scores": {
        "style": 1-5,
        "structure": 1-5,
        "alignment": 1-5
    },
    "reason": "Comprehensive judgment rationale based on analysis",
    "overall": 1-5
}
```

**Notes**:
- The three scores in `scores` correspond to the similarity ratings for each dimension
- `overall` is the comprehensive score, not necessarily a simple average of the three scores
\end{Prompt}

\section{Action Graph Similarity: Construction and Sub-metrics}
\label{app:ags}

AGS starts from an Action Flow Graph built from the tool trajectory and then extracts three sub-metrics from that graph.

\subsection{Graph Construction}
\label{app:graph}

Given a dialogue trajectory $\tau$, we construct an Action Flow Graph $G = (V, E_s, E_d)$ based on the tool call sequence. Each node $v \in V$ represents a tool call with attributes $v = (\text{name}, \text{args}, \text{result})$.

\textbf{Dependency Edge Detection.}
\label{app:dep-prompt}
A dependency edge $(u, v) \in E_d$ is established when an argument
value of $v$ is derived from the result of $u$. Detection follows
a two-stage process. First, leaf values are recursively extracted
from the destination tool's arguments and matched against previous
tool results. Values shorter than three characters, empty values,
and a blacklist of non-informative tokens (e.g., boolean markers,
status words, placeholders) are discarded. Surviving matches are
then verified by an LLM judge for semantic validity. The prompt
template is shown below.

\begin{Prompt}[label={prompt:dep},title={Prompt for Dependency Edge Detection}]
    
# Role
You are an expert in analyzing data-flow dependencies between tool calls in agent trajectories.

# Task
Given a candidate dependency edge between two tool calls and a matched value, determine whether the matched value in the destination tool's arguments was actually derived from the source tool's result.

# Judgment Criteria
A TRUE dependency holds when:
- The matched value first becomes available through the source tool's result
- The destination tool uses this value as an input that could not have been known prior to the source call

A FALSE dependency (spurious match) holds when:
- The value was already available from user input or system context before the source tool was called
- The match is coincidental (e.g., common dates, generic status codes, or short numeric values that appear frequently across tool results)

# Input
## Source Tool Call
- Tool: {src_tool}
- Result: {src_result}

## Destination Tool Call
- Tool: {dst_tool}
- Arguments: {dst_args}

## Matched Value
"{matched_value}"

# Output Format
```
{
    "is_true_dependency": true/false,
    "reasoning": "One sentence explaining why the value was or was not derived from the source result"
}
```
\end{Prompt}

\textbf{Sequential Edge Construction.} A sequential edge $(u, v) \in E_s$ connects temporally adjacent tool calls according to their execution order. Sequential edges are added only when the target node has no incoming dependency edges, ensuring that data dependencies take precedence over temporal ordering.

\subsection{Optional Tool Agreement}\label{app:node}

Optional Tool Agreement $S_{\text{node}}$ measures the consistency of optional tool choices between two models.

\textbf{Tool Classification.} For each task $t$, let $\mathcal{M}_t^*$ denote the set of models that successfully complete the task. Mandatory tools are those invoked by all successful models:
$$\mathcal{F}_t^{\text{mandatory}} = \bigcap_{M \in \mathcal{M}_t^*} \text{Tools}(M, t)$$
Optional tools are those appearing in some but not all successful trajectories:
$$\mathcal{F}_t^{\text{opt}} = \bigcup_{M \in \mathcal{M}_t^*} \text{Tools}(M, t) \;\setminus\; \mathcal{F}_t^{\text{mandatory}}$$

\textbf{Agreement Computation.} For each task $t$, we count how many optional tools receive the same decision from both models (both invoke it or both skip it), and divide by the total number of optional tools. $S_{\text{node}}$ is the average of this ratio across all tasks. Tasks where $\mathcal{F}_t^{\text{opt}} = \emptyset$ are excluded.

\subsection{Sequential Pattern Similarity}
\label{app:seq}

Sequential Pattern Similarity $S_{\text{seq}}$ measures whether two models exhibit similar local execution habits between adjacent tool calls. Each trajectory is represented as a three-dimensional feature vector.

\textbf{Tool Type Classification.} We classify tools into read and write operations based on their side effects. Read tools query information without modifying system state; write tools perform modifications. The classification follows the official $\tau$-bench and $\tau^2$-bench tool definitions (see Table~\ref{tab:tool-classification}).

\textbf{Feature Definitions.} Let $n_w$ denote the number of write operations in the trajectory and $n_{\text{err}}$ the number of tool calls that return an error response. The three features are:
\begin{itemize}[noitemsep, topsep=0pt]
\item \textit{Post-write verification rate} $r_{\text{verify}} = n_{\text{w} \to \text{r}} / n_w$, where $n_{\text{w} \to \text{r}}$ is the number of write operations immediately followed by a read operation.
\item \textit{Pre-write confirmation rate} $r_{\text{confirm}} = n_{\text{r} \to \text{w}} / n_w$, where $n_{\text{r} \to \text{w}}$ is the number of write operations immediately preceded by a read operation.
\item \textit{Error retry rate} $r_{\text{retry}} = n_{\text{retry}} / n_{\text{err}}$, where $n_{\text{retry}}$ is the number of error responses followed by retrying the same tool. Errors are detected by keyword matching against the tool result text.
\end{itemize}

\textbf{Similarity Computation.} Each trajectory yields a
three-dimensional feature vector $(r_{\text{verify}},
r_{\text{confirm}}, r_{\text{retry}})$. For each task, we compute
the cosine similarity between the two models' vectors and average
across all tasks to obtain $S_{\text{seq}}$. When both vectors are
zero on the same task (e.g., neither trajectory contains any write
operations), cosine similarity is undefined; we set it to 1.0,
since both models behave identically on that task. When exactly one
vector is zero, we set the similarity to 0.0.

\begin{table*}[t]
\centering
\caption{Read, write, and generic tool classification for the airline, retail, and telecom domains from $\tau$-Bench and $\tau^2$-Bench. This classification is used to compute $S_{\text{seq}}$.}
\label{tab:tool-classification}
\small
\begin{tabular}{@{}llp{12cm}@{}}
\toprule
\textbf{Domain} & \textbf{Type} & \textbf{Tools} \\
\midrule
\multirow{2}{*}{Airline}
& Read & get\_reservation\_details, get\_user\_details, list\_all\_airports, search\_direct\_flight, search\_onestop\_flight \\
& Write & book\_reservation, cancel\_reservation, send\_certificate, update\_reservation\_baggages, update\_reservation\_flights, update\_reservation\_passengers \\
\midrule
\multirow{2}{*}{Retail}
& Read & find\_user\_id\_by\_email, find\_user\_id\_by\_name\_zip, get\_order\_details, get\_product\_details, get\_user\_details, list\_all\_product\_types \\
& Write & cancel\_pending\_order, exchange\_delivered\_order\_items, modify\_pending\_order\_address, modify\_pending\_order\_items, modify\_pending\_order\_payment, modify\_user\_address, return\_delivered\_order\_items \\
\midrule
\multirow{2}{*}{Telecom}
& Read & get\_customer\_by\_phone, get\_customer\_by\_id, get\_customer\_by\_name, get\_details\_by\_id, get\_bills\_for\_customer, get\_data\_usage \\
& Write & suspend\_line, resume\_line, send\_payment\_request, enable\_roaming, disable\_roaming, refuel\_data \\
\midrule
All & Generic & calculate, transfer\_to\_human\_agents \\
\bottomrule
\end{tabular}
\end{table*}

\subsection{Dependency Pattern Similarity}\label{app:dep}

Dependency Pattern Similarity $S_{\text{dep}}$ measures whether two models exhibit similar patterns in reusing tool outputs. Each trajectory is represented as a three-dimensional feature vector based on the dependency-edge subgraph $(V, E_d)$ of the Action Flow Graph.

\textbf{Feature Definitions.} In a tool-use trajectory, some tool calls receive arguments from the results of earlier calls (connected by dependency edges), while others take values directly from user input. The three features below characterize how much and how broadly a model relies on inter-tool passing. Let $n_{\text{in}}$ denote the number of nodes with at least one incoming dependency edge, $n_{\text{out}}$ the number of nodes with at least one outgoing dependency edge, and $n_{\text{fan}}$ the number of nodes with two or more outgoing dependency edges. The three features are:
\begin{itemize}[noitemsep, topsep=0pt]
\item \textit{Output reuse rate} $r_{\text{reuse}} = n_{\text{in}} / (|V| - 1)$, the fraction of tool calls (excluding the first) whose inputs come from a preceding call's output.
\item \textit{Longest dependency chain} $d_{\max}$, the length of the longest directed path in $(V, E_d)$.
\item \textit{Output fan-out rate} $r_{\text{fanout}} = n_{\text{fan}} / n_{\text{out}}$, the fraction of tool outputs reused by two or more subsequent calls.
\end{itemize}

\textbf{Similarity Computation.} Each trajectory yields a
three-dimensional feature vector $(r_{\text{reuse}}, d_{\max},
r_{\text{fanout}})$. The final score is the per-task cosine
similarity averaged across all tasks, with the same zero-vector
convention as $S_{\text{seq}}$.

\subsection{Benchmark Tool Definitions}
\label{app:tools}

The tool benchmarks come from $\tau$-bench~\citep{tau-bench} and $\tau^2$-bench~\citep{tau2-bench}, spanning airline, retail, and telecom tasks. Table~\ref{tab:tool-classification} lists the read/write/generic split used for $S_{\text{seq}}$.

\subsubsection{Tool Signatures and Examples}
\label{app:tool-examples}

The examples below illustrate the signatures and invocation patterns covered by the metrics.

\textbf{Airline Domain.} The airline domain involves flight booking, reservation management, and customer service operations.

\begin{itemize}[noitemsep, topsep=0pt]
\item get\_user\_details(user\_id) $\rightarrow$ Returns user profile including name, email, date of birth, payment methods, and membership tier.
\item search\_direct\_flight(origin, destination, date) $\rightarrow$ Returns available direct flights with flight numbers, departure/arrival times, and prices.
\item update\_reservation\_flights(reservation\_id, cabin, flights, payment\_id) $\rightarrow$ Modifies flight selection for an existing reservation.
\end{itemize}

\textbf{Retail Domain.} The retail domain covers e-commerce operations including order management, returns, and customer account updates.

\begin{itemize}[noitemsep, topsep=0pt]
\item find\_user\_id\_by\_email(email) $\rightarrow$ Returns user ID for the given email address.
\item get\_order\_details(order\_id) $\rightarrow$ Returns order information including items, status, shipping address, and payment method.
\item exchange\_delivered\_order\_items(order\_id, item\_ids, new\_item\_ids, payment\_method\_id) $\rightarrow$ Processes item exchange for a delivered order.
\end{itemize}

\textbf{Telecom Domain.} The telecom domain handles mobile service management including line operations, billing, and data plans.

\begin{itemize}[noitemsep, topsep=0pt]
\item get\_customer\_by\_phone(phone\_number) $\rightarrow$ Returns customer profile and associated phone lines.
\item get\_data\_usage(customer\_id, line\_id) $\rightarrow$ Returns current data usage statistics for a specific line.
\item refuel\_data(customer\_id, line\_id, gb\_amount) $\rightarrow$ Adds additional data to a phone line.
\end{itemize}

\section{Ablation Studies}

We ablate two design choices (stage alignment in RPS and mandatory/optional tool separation in AGS) and report the full sub-metric breakdown of the controlled distillation experiment from Section~\ref{subsec:lineage}.

\subsection{Stage Alignment Ablation}\label{app:stage-ablation}

We compare RPS with and without stage alignment using Claude Sonnet 4.5 (thinking) as the reference model. For each model, we evaluate 15 $\tau$-bench tasks and run the judge three times per case, then report the within-case score standard deviation ($\sigma_{\text{within}}$).

\begin{table}[t]
\centering
\caption{Within-case RPS score standard deviation ($\sigma_{\text{within}}$) with and without stage alignment for three models compared against Claude Sonnet 4.5 (thinking) on 15 $\tau$-bench tasks, using three judge runs per case.}
\label{tab:stage-ablation}
\small
\resizebox{0.85\columnwidth}{!}{%
\begin{tabular}{@{}lccc@{}}
\toprule
\textbf{Model} & \textbf{With alignment} & \textbf{Without alignment} & \textbf{Reduction} \\
\midrule
GPT-4.1 & 0.248 & 0.446 & $-$44\% \\
Kimi-K2 (thinking) & 0.233 & 0.458 & $-$49\% \\
DeepSeek-R1 & 0.255 & 0.563 & $-$55\% \\
\bottomrule
\end{tabular}%
}
\end{table}

Without stage alignment, $\sigma_{\text{within}}$ increases from 0.23--0.26 to 0.45--0.56. The reduction from alignment ranges from 44\% to 55\%.

\subsection{Mandatory/Optional Tool Separation Ablation}\label{app:mandatory-ablation}

We compare $S_{\text{node}}$ computed on optional tools only versus all tools, using Claude Sonnet 4.5 (thinking) as the reference model.

\begin{table}[t]
\centering
\caption{$S_{\text{node}}$ computed on optional tools only versus all tools, using Claude Sonnet 4.5 (thinking) as the reference model.}
\label{tab:mandatory-ablation}
\small
\resizebox{0.75\columnwidth}{!}{%
\begin{tabular}{@{}lccc@{}}
\toprule
\textbf{Model} & \textbf{Optional only} & \textbf{All tools} & \textbf{$\Delta$} \\
\midrule
Gemini 3 Pro & 66.7\% & 82.5\% & +15.8 pp \\
DeepSeek-R1 & 69.4\% & 79.7\% & +10.4 pp \\
DeepSeek-V3-0324 & 75.6\% & 86.0\% & +10.4 pp \\
Claude Opus 4.1 (thinking) & 87.5\% & 87.5\% & 0.0 pp \\
\bottomrule
\end{tabular}%
}
\end{table}

Using all tools raises non-family models by 12.2 pp on average. The gap between Gemini 3 Pro and Claude Opus 4.1 (thinking) drops from 20.8 pp to 5.0 pp. Claude Opus 4.1 (thinking) is unchanged, so its higher similarity comes from optional-tool choices rather than mandatory overlap.

\subsection{Controlled Distillation: Sub-metric Analysis}\label{app:distill-analysis}

Table~\ref{tab:distill-full} reports the full sub-metric breakdown for the controlled distillation experiment in Section~\ref{subsec:lineage}.

\begin{table}[t]
\centering
\caption{Full controlled-distillation results for LoRA fine-tuning of Qwen2.5-14B-Instruct on trajectories generated by Claude Sonnet 4.5 (thinking).}
\label{tab:distill-full}
\small
\resizebox{\columnwidth}{!}{%
\begin{tabular}{@{}lcccccc@{}}
\toprule
\textbf{Metric} & \textbf{Base$\rightarrow$Claude} & \textbf{Dist$\rightarrow$Claude} & \textbf{$\Delta$(Teacher)} & \textbf{Base$\rightarrow$R1} & \textbf{Dist$\rightarrow$R1} & \textbf{$\Delta$(Control)} \\
\midrule
$S_{\text{node}}$ & 0.55 & 0.60 & +0.05 & 0.58 & 0.41 & $-$0.17 \\
$S_{\text{seq}}$ & 0.50 & 0.58 & +0.08 & 0.63 & 0.49 & $-$0.14 \\
$S_{\text{dep}}$ & 0.73 & 0.99 & +0.27 & 0.71 & 0.87 & +0.16 \\
AGS & 0.59 & 0.72 & +0.13 & 0.64 & 0.59 & $-$0.05 \\
RPS & 2.03 & 3.39 & +1.36 & 2.00 & 2.81 & +0.81 \\
GED & 0.42 & 0.65 & +0.23 & 0.39 & 0.59 & +0.20 \\
\bottomrule
\end{tabular}%
}
\end{table}

Transfer is uneven across behavioral dimensions. $S_{\text{dep}}$ rises to 0.99 (+0.27), replicating nearly all teacher dependency patterns. $S_{\text{node}}$ increases by only +0.05, while $S_{\text{seq}}$ shows a moderate gain of +0.08.

RPS rises toward both the teacher (+1.36) and the control (+0.81). Fine-tuning on high-quality trajectory data improves response fluency in general, which raises RPS against both references. The teacher-control gap remains +0.55.

GED rises by +0.23 toward the teacher and +0.20 toward the control. The teacher-control gap for GED is 0.03, compared with 0.18 for AGS.

\subsection{Controlled Distillation: Training Details}\label{app:distill-setup}

We fine-tune Qwen2.5-14B-Instruct on 200 $\tau$-bench trajectories generated by Claude Sonnet 4.5 (thinking) using LoRA. The trajectories cover the two $\tau$-Bench domains (airline and retail). Training hyperparameters are 10 epochs, learning rate 2e-5, and LoRA rank 16. DeepSeek R1 serves as the non-teacher control, and no R1 trajectories are used during training.

\section{Robustness and Sensitivity Analysis}
\label{app:robustness}

We test metric stability under changes to the model pool, reference model, and sampling settings.

\subsection{Sensitivity Analysis of \texorpdfstring{$S_{\text{node}}$}{S node}}
\label{app:bootstrap}

The mandatory/optional split depends on the set of models included in the analysis. We test its stability on the same 50-task subset used in Section~\ref{subsec:lineage} by sampling 100 subsets, each formed by removing two models other than the reference and target models, and recomputing $S_{\text{node}}$.

\begin{table}[t]
\centering
\caption{Sensitivity analysis of $S_{\text{node}}$ on the 50-task lineage subset with Claude Sonnet 4.5 (thinking) as the reference model, using 100 resampled model pools formed by removing two non-reference, non-target models. CV denotes coefficient of variation.}
\label{tab:bootstrap}
\small
\resizebox{0.85\columnwidth}{!}{%
\begin{tabular}{@{}lccc@{}}
\toprule
\textbf{Model} & \textbf{Full $S_{\text{node}}$} & \textbf{Mean $\pm$ Std} & \textbf{CV} \\
\midrule
Kimi-K2 (thinking) & 82.6\% & 80.8\% $\pm$ 1.2\% & 1.5\% \\
GLM-4.6 & 79.9\% & 78.4\% $\pm$ 1.2\% & 1.6\% \\
Kimi-K2 & 73.8\% & 72.1\% $\pm$ 1.8\% & 2.5\% \\
DeepSeek-V3.1 (thinking) & 72.2\% & 70.6\% $\pm$ 1.5\% & 2.2\% \\
DeepSeek-R1 & 71.0\% & 69.6\% $\pm$ 1.3\% & 1.9\% \\
GPT-5 & 69.4\% & 67.9\% $\pm$ 1.9\% & 2.7\% \\
Gemini-2.5-Pro & 68.2\% & 66.1\% $\pm$ 1.9\% & 2.8\% \\
Doubao-1.6 (high) & 63.0\% & 61.0\% $\pm$ 1.0\% & 1.6\% \\
\bottomrule
\end{tabular}%
}
\end{table}

Across model pairs, CV ranges from 1.5\% to 2.8\%. The ranking remains stable across resampled model pools.

\subsection{Multi-Reference Analysis}
\label{app:multi-ref}

We test whether the metrics capture directional similarity rather than benchmark-level convergence. AGS is computed on the same 50-task subset used in Section~\ref{subsec:lineage}, once with GPT-5 as the reference model and once with Claude Sonnet 4.5 (thinking) as the reference model.

\begin{table}[t]
\centering
\caption{AGS on the 50-task lineage subset using Claude Sonnet 4.5 (thinking) or GPT-5 as the reference model. Positive $\Delta$ indicates higher similarity to Claude.}
\label{tab:multi-ref}
\small
\resizebox{0.75\columnwidth}{!}{%
\begin{tabular}{@{}lccc@{}}
\toprule
\textbf{Model} & \textbf{vs Claude} & \textbf{vs GPT-5} & \textbf{$\Delta$} \\
\midrule
Kimi-K2 (thinking) & 81.9\% & 76.1\% & +5.8\% \\
DeepSeek-R1 & 77.9\% & 70.1\% & +7.7\% \\
GLM-4.6 & 79.6\% & 76.5\% & +3.1\% \\
DeepSeek-V3.1 (thinking) & 75.9\% & 73.9\% & +2.0\% \\
Kimi-K2 & 75.9\% & 74.9\% & +1.0\% \\
Doubao-1.6 (high) & 72.0\% & 67.5\% & +4.5\% \\
Gemini-2.5-Pro & 71.2\% & 67.9\% & +3.3\% \\
\midrule
Claude-Opus-4.1 (thinking) & 82.1\% & 73.5\% & +8.6\% \\
\bottomrule
\end{tabular}%
}
\end{table}

All evaluated models show higher AGS with Claude than with GPT-5. Kimi-K2 (thinking) is 5.8\% higher against Claude, while Claude Opus 4.1 (thinking) shows the largest gap at 8.6\%. The metric therefore changes with the reference model instead of collapsing to a benchmark-specific similarity baseline.

\subsection{Dependency Edge Detection Accuracy}
\label{app:dep-precision}

We randomly sample 50 edges identified by the LLM judge as dependencies and compare them against human annotations.

DeepSeek-V3 reaches 96\% accuracy on this sample. The primary error source is ambiguous values that could plausibly originate from either user input or a previous tool result.

\subsection{RPS Judge Consistency}
\label{app:rps-consistency}

We measure agreement between two judges, Gemini-2.5-Flash-Thinking and DeepSeek-V3, on the same set of 50 airline-domain stage comparisons for Claude Sonnet 4.5 (thinking) versus Kimi-K2 (thinking).

\begin{table}[t]
\centering
\caption{Inter-rater agreement for RPS scoring between Gemini-2.5-Flash-Thinking and DeepSeek-V3 on 50 airline-domain stage comparisons for Claude Sonnet 4.5 (thinking) versus Kimi-K2 (thinking).}
\label{tab:rps-consistency}
\small
\resizebox{0.6\columnwidth}{!}{%
\begin{tabular}{@{}lc@{}}
\toprule
\textbf{Metric} & \textbf{Value} \\
\midrule
Sample Size & 50 \\
Cohen's Kappa (quadratic) & 0.70 \\
Exact Agreement & 46\% \\
Close Agreement ($\pm 1$) & 94\% \\
\bottomrule
\end{tabular}%
}
\end{table}

Quadratic-weighted Cohen's $\kappa$ is 0.70. Exact agreement is 46\%, and agreement within one point is 94\%.

\label{app:judge-temp}
\textbf{Ranking stability across judge temperatures.} We vary the temperature of Gemini-2.5-Flash-Thinking across t=0.0, 0.7, and 1.0 on 100 randomly sampled retail-domain intent blocks, evaluating three model pairs against Claude Sonnet 4.5 (thinking).

\begin{table}[t]
\centering
\caption{RPS scores from Gemini-2.5-Flash-Thinking at three judge temperatures on 100 randomly sampled retail-domain intent blocks, using Claude Sonnet 4.5 (thinking) as the reference model.}
\label{tab:judge-temp}
\small
\resizebox{0.75\columnwidth}{!}{%
\begin{tabular}{@{}lcccc@{}}
\toprule
\textbf{Model} & \textbf{t=0.0} & \textbf{t=0.7} & \textbf{t=1.0} & \textbf{Range} \\
\midrule
Kimi-K2 (thinking) & 3.708 & 3.433 & 3.505 & 0.275 \\
GPT-4.1 & 3.351 & 3.273 & 3.144 & 0.207 \\
DeepSeek-V3-0324 & 2.978 & 2.840 & 2.876 & 0.138 \\
\bottomrule
\end{tabular}%
}
\end{table}

The ranking Kimi-K2 > GPT-4.1 > DeepSeek-V3-0324 is unchanged at all three temperatures. Absolute score ranges span 0.138--0.275, while all pairwise gaps remain positive.

\textbf{Test-retest reliability.} We rerun the same judge twice at each temperature on the same 100 intent blocks for Kimi-K2 (thinking) versus Claude Sonnet 4.5 (thinking).

\begin{table}[t]
\centering
\caption{Test-retest reliability of Gemini-2.5-Flash-Thinking across judge temperatures on 100 intent blocks for Kimi-K2 (thinking) versus Claude Sonnet 4.5 (thinking).}
\label{tab:judge-retest}
\small
\resizebox{0.7\columnwidth}{!}{%
\begin{tabular}{@{}lccc@{}}
\toprule
\textbf{Temperature} & \textbf{Pearson $r$} & \textbf{Exact} & \textbf{$\pm$1} \\
\midrule
0.0 & 0.852 & 79.3\% & 94.6\% \\
0.7 & 0.727 & 53.8\% & 93.5\% \\
1.0 & 0.590 & 39.8\% & 88.2\% \\
\bottomrule
\end{tabular}%
}
\end{table}

All three reliability metrics decrease monotonically with judge temperature. At the default setting (t=0.7), exact agreement falls to 53.8\%, while $\pm$1 agreement remains 93.5\%.

\subsection{Temperature Sensitivity}\label{app:temp}

We vary the sampling temperature of the compared models while keeping Claude Sonnet 4.5 (thinking) as the reference model.

\begin{table}[t]
\centering
\caption{AGS across model sampling temperatures for Kimi-K2 (thinking) and DeepSeek-V3-0324 on 50 randomly sampled $\tau$-bench tasks, with three runs per temperature and Claude Sonnet 4.5 (thinking) as the reference model.}
\label{tab:temp}
\small
\resizebox{\columnwidth}{!}{%
\begin{tabular}{@{}llcccc@{}}
\toprule
\textbf{Model} & \textbf{Temperature} & \textbf{AGS} & $S_{\text{node}}$ & $S_{\text{seq}}$ & $S_{\text{dep}}$ \\
\midrule
\multirow{4}{*}{Kimi-K2 (thinking)}
& 0.3 & 0.862$\pm$0.049 & 0.880 & 0.754 & 0.996 \\
& 0.7 & 0.836$\pm$0.010 & 0.846 & 0.746 & 0.954 \\
& 1.0 & 0.809$\pm$0.026 & 0.835 & 0.729 & 0.932 \\
& Range & 0.053 & 0.045 & 0.025 & 0.064 \\
\midrule
\multirow{4}{*}{DeepSeek-V3-0324}
& 0.3 & 0.675$\pm$0.043 & 0.720 & 0.529 & 0.764 \\
& 0.7 & 0.649$\pm$0.025 & 0.693 & 0.540 & 0.734 \\
& 1.0 & 0.687$\pm$0.051 & 0.706 & 0.536 & 0.750 \\
& Range & 0.038 & 0.027 & 0.011 & 0.030 \\
\bottomrule
\end{tabular}%
}
\end{table}

Intra-model AGS variation is 0.053 for Kimi-K2 (thinking) and 0.038 for DeepSeek-V3-0324. Across all temperature settings, Kimi-K2 (thinking) remains at AGS $\approx$ 0.81--0.86, while DeepSeek-V3-0324 remains at AGS $\approx$ 0.65--0.69. Even under the most adversarial comparison (Kimi at t=1.0 vs.\ DeepSeek at t=0.3), the inter-model gap (0.134) exceeds the maximum intra-model range by over 2.5$\times$. Temperature-induced variation is also less than half of the distillation signal magnitude (13 pp, Table~\ref{tab:distill-valid}).

\subsection{Correlation with Baseline Metrics}\label{app:correlation}

We compute pairwise correlations on 16 non-Anthropic models evaluated on $\tau$-bench with Claude Sonnet 4.5 (thinking) as the reference model.

\begin{table}[t]
\centering
\caption{Pairwise correlations among GED, AGS, RSE, and RPS over 16 non-Anthropic models evaluated against Claude Sonnet 4.5 (thinking) on $\tau$-bench.}
\label{tab:correlation}
\small
\resizebox{0.75\columnwidth}{!}{%
\begin{tabular}{@{}lcccc@{}}
\toprule
\textbf{Metric Pair} & \textbf{Pearson $r$} & \textbf{$p$-val} & \textbf{Spearman $\rho$} & \textbf{$p$-val} \\
\midrule
GED vs AGS & 0.806 & <0.001 & 0.844 & <0.001 \\
RSE vs RPS & 0.682 & 0.004 & 0.701 & 0.003 \\
AGS vs RPS & 0.491 & 0.054 & 0.484 & 0.057 \\
\bottomrule
\end{tabular}%
}
\end{table}

GED--AGS and RSE--RPS show moderate-to-high correlation. AGS--RPS is lower ($r$=0.491, $p$=0.054), consistent with the two metrics targeting different behavioral streams.

\textbf{Measurement resolution.} RSE places 12 of the 16 models within a 0.51-point band (2.05--2.56 on a 1--5 scale), while RPS spans 1.25 points (2.40--3.65) and adds Style, Structure, and Alignment sub-scores.

\textbf{Interpretability.} GED is a single scalar. For Kimi-K2 (thinking), GED = 78.1, whereas AGS decomposes similarity into $S_{\text{dep}}$ = 94.7\%, $S_{\text{node}}$ = 82.6\%, and $S_{\text{seq}}$ = 70.8\%, separating dependency reuse from sequential habits.

\subsection{Framework Choice Analysis}\label{app:framework}

As a preliminary analysis, we examine whether the agent framework affects behavioral similarity. We compare a generic ReAct framework with minimal behavioral constraints against OpenHands, which adds explicit instructions such as ``combine multiple actions'' and ``use efficient tools''. The comparison uses GPT-5 and Claude Sonnet 4.5 (thinking) on 14 $\tau$-bench tasks.

\begin{table}[t]
\centering
\caption{AGS between GPT-5 and Claude Sonnet 4.5 (thinking) under a baseline ReAct framework and OpenHands (preliminary, 14 tasks).}
\label{tab:framework}
\small
\resizebox{0.65\columnwidth}{!}{%
\begin{tabular}{@{}lccc@{}}
\toprule
\textbf{Sub-metric} & \textbf{Baseline} & \textbf{OpenHands} & \textbf{$\Delta$} \\
\midrule
$S_{\text{node}}$ & 0.21 & 0.76 & +0.55 \\
$S_{\text{seq}}$ & 0.76 & 0.89 & +0.13 \\
$S_{\text{dep}}$ & 0.99 & 1.00 & +0.01 \\
\bottomrule
\end{tabular}%
}
\end{table}
\begin{figure*}[ht]
    \centering
    \includegraphics[width=\textwidth]{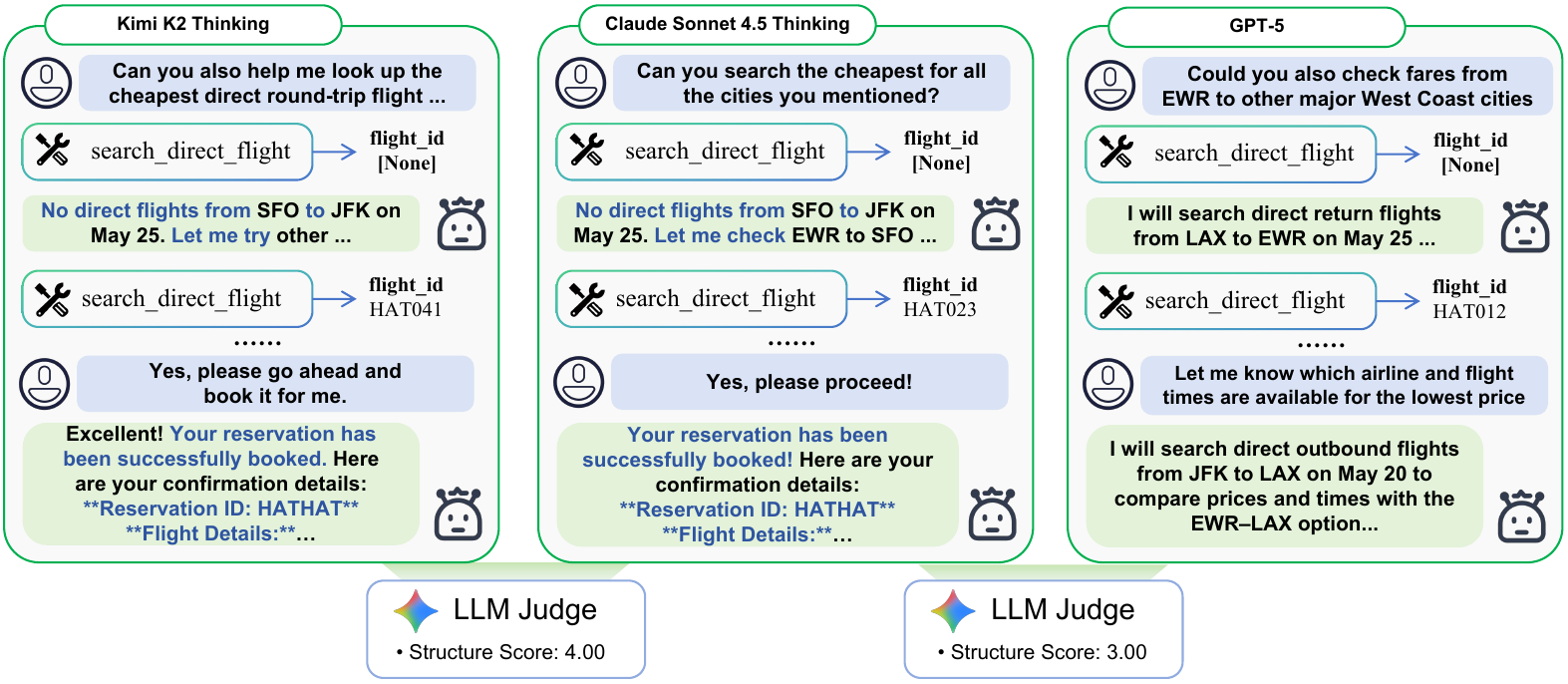}
    \caption{RPS \textbf{structure} comparison on an airline reservation task. Kimi-K2 (thinking) and Claude Sonnet 4.5 (thinking) converge in response formatting; GPT-5 uses a more compact style.}
    \label{fig:case-appdx1}
\end{figure*}

$S_{\text{node}}$ rises by +0.55 under OpenHands, whereas $S_{\text{seq}}$ and $S_{\text{dep}}$ shift by only +0.13 and +0.01. Framework constraints therefore affect tool selection more than execution order or data-flow reuse. This is why we use a generic ReAct framework with minimal constraints in the main experiments.

\section{Additional Case Studies}
\label{app:case-studies}

\subsection{RPS Case Studies}
\label{app:rps-case}

We provide additional case studies for the structure and alignment dimensions of RPS.

\textbf{Structural convergence can be a discriminative feature across models.} Figure~\ref{fig:case-appdx1} shows a case from an airline reservation task. Claude Sonnet 4.5 (thinking) and Kimi-K2 converge spontaneously in response structure. When the initial search fails, both models adopt nearly identical phrasing to describe the fallback strategy. In the final response stage, both shift to a highly organized, template-like format with bold headings and list-based presentation, which neither the user request nor the system prompt requires. These structural similarities contribute to higher RPS scores. GPT-5 instead uses a more directive and compact structure with limited formatting, which leads to a lower RPS score.

\begin{figure*}[ht]
    \centering
    \includegraphics[width=\textwidth]{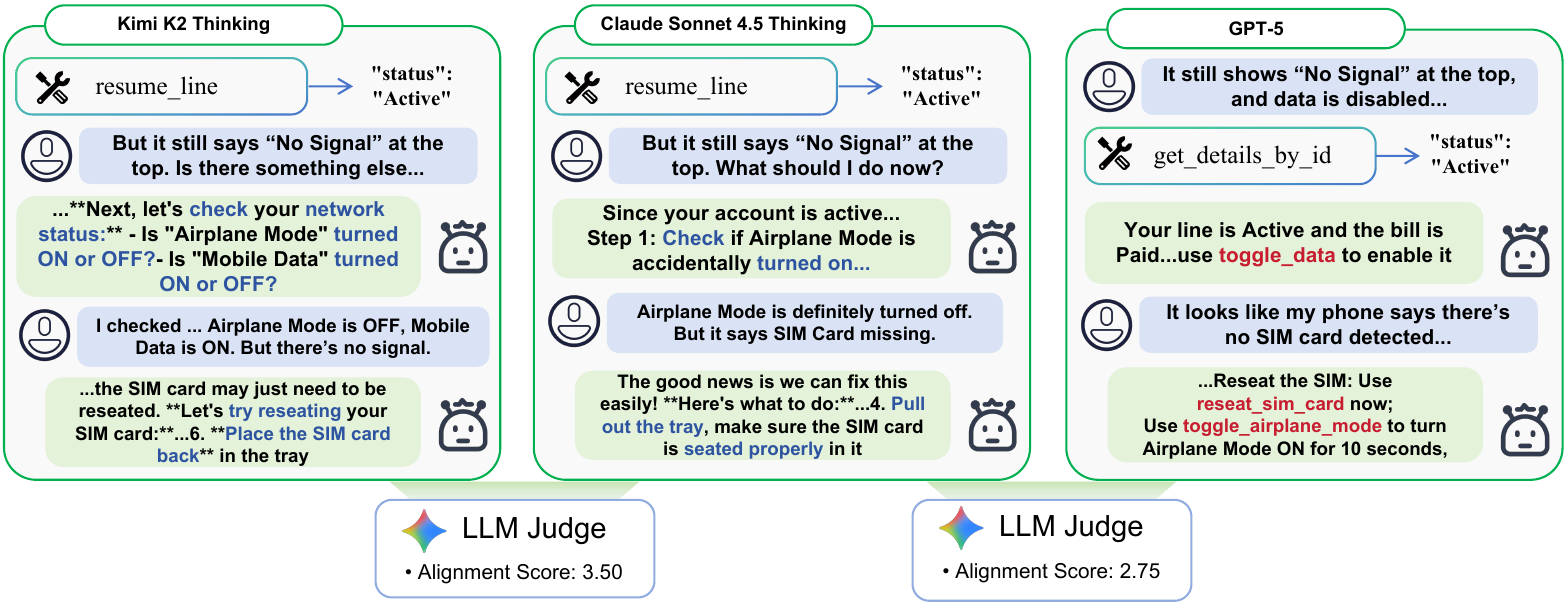}
    \caption{RPS \textbf{alignment} comparison on a telecom troubleshooting task. GPT-5 hallucinates agent-side tool instructions (marked in red); Kimi-K2 (thinking) and Claude Sonnet 4.5 (thinking) provide real device instructions instead.}
    \label{fig:case-appdx2}
\end{figure*}

\textbf{Alignment pattern similarity highlights hallucinations in interaction.} In the telecom domain, models differ in how they interpret tool boundaries, revealing alignment patterns that serve as indicators of behavioral similarity. In Figure~\ref{fig:case-appdx2}, GPT-5 instructs the user to invoke agent-side simulation tools, erroneously assuming user access to these functions. Claude Sonnet 4.5 (thinking) and Kimi-K2 instead recognize this limitation and provide step-by-step instructions for real device operations. Their wording differs, but the troubleshooting strategy is aligned. This yields higher RPS scores (around 3.75), whereas GPT-5 receives 1.50 because of the hallucinated tool guidance. Even when surface phrasing differs, the alignment sub-metric still amplifies discrepancies caused by output hallucinations.

\subsection{AGS Case Studies}
\label{app:ags-case}

We provide detailed analysis of $S_{\text{seq}}$ and $S_{\text{dep}}$ sub-metrics.

\textbf{Sequential Pattern Similarity captures consistency in tool invocation order.}
In the both-correct setting (Table~\ref{tab:ags_submetrics}), all three pairs have high and similar $S_{\text{seq}}$ values: 0.912 for Kimi--GPT, 0.904 for GPT--Claude, and 0.892 for Kimi--Claude. When all models succeed, they often follow similar correct paths, so sequential similarity is less discriminative. In the both-wrong setting, however, the Kimi--Claude pair reaches 0.829, compared with 0.680 for GPT--Claude and 0.583 for Kimi--GPT. Once tasks fail, retry order and recovery strategy become model-specific decisions. The high similarity between Kimi and Claude in this setting indicates shared error-handling patterns. Notably, GED ranks the Kimi--GPT pair highest in the both-wrong setting at 0.594, while the Kimi--Claude pair scores only 0.558. $S_{\text{seq}}$ captures sequential-level convergence that GED misses.

\textbf{Dependency Pattern Similarity captures consistency in inter-tool parameter dependencies.}
Across all three settings, the Kimi--Claude pair has the highest $S_{\text{dep}}$, ranging from 0.917 in the mixed setting to 0.993 in the both-correct setting, while the other two pairs remain below 0.90. The gap is widest in the both-correct setting, where the next-highest pair reaches only 0.882. Kimi and Claude therefore align not only in tool choice but also in how outputs from one tool feed into later calls. As with $S_{\text{seq}}$, GED ranks the Kimi--GPT pair highest in the both-wrong setting, while $S_{\text{dep}}$ ranks the Kimi--Claude pair highest. $S_{\text{dep}}$ therefore captures dependency-level consistency that GED misses.

\end{document}